\documentclass[review]{elsarticle}

\newdefinition{rmk}{Remark}

\usepackage{hyperref}
\usepackage{amsmath,amssymb}
\usepackage{float}
\usepackage{caption}
\usepackage{subfigure}
\usepackage{rotating}
\usepackage{tabularx}
\usepackage{booktabs}

\usepackage{xcolor} % for comments ...
\usepackage{verbatim}

\usepackage{graphicx}
\usepackage{amsmath}
 \usepackage{amssymb} %  for mathbb
 
\usepackage{tikz} % for tikzpicture
\usepackage{pgfplots}
\pgfplotsset{width=10cm,compat=1.15}
\usepackage{mathrsfs}
\usetikzlibrary{arrows}
\usetikzlibrary{intersections}
\usetikzlibrary{fillbetween}
\usetikzlibrary{calc}
 
 \usepackage{verbatim} % for 'comment'
 \usepackage{booktabs} % for toprule, etc
 
 \usepackage{xcolor} % for colored comments
 
 \usepackage{rotating} % zum Drehen der Tabelle
 \usepackage{url} % for links
 \usepackage{subfigure} % for using subfigures
 \usepackage[linesnumbered,ruled, noend]{algorithm2e} % to display algorithms
\graphicspath{{./figures/}} 

\newcommand{\R}{\mathbb{R}}

\DeclareMathOperator{\MSE}{MSE}
\DeclareMathOperator{\MSEU}{\MSE_\mathcal{U}}
\DeclareMathOperator{\MSEF}{\MSE_\mathcal{F}}

\DeclareMathOperator{\PINNpW}{\mathcal{PINN}_p^{W}}
\newcommand{\dt}[1]{\frac{\mathrm{d} #1}{\mathrm{d}t}}

\def\new#1{{{#1}}}

\date{December 11, 2022}

\begin{document}

\begin{frontmatter}
%\title{Optimized PINNs}
%\title{PINN Training using Multiobjective Optimization: The Trade-off between Data Loss and Residual Loss}
\title{PINN Training using Biobjective Optimization: \\
The Trade-off between Data Loss and Residual Loss}
%\subtitle{}

\author[uniwo]{Fabian Heldmann} 
 \ead{heldmann@uni-wuppertal.de} 

\author[uniw]{Sarah \new{Berkhahn}}
\ead{streibert@uni-wuppertal.de}        

\author[uniw]{Matthias Ehrhardt}
\ead{ehrhardt@uni-wuppertal.de}

\author[uniwo]{Kathrin Klamroth}
 \ead{klamroth@uni-wuppertal.de} 

\address[uniw]{University of Wuppertal, 
%Chair of Applied Mathematics and Numerical Analysis,
\new{Chair of Applied and Computational Mathematics},
Gau\ss strasse~20, 42119 Wuppertal, Germany}

\address[uniwo]{University of Wuppertal, Chair of Optimization, Gau\ss strasse~20, 42119 Wuppertal, Germany}

%%%%%%%%%%%%%%%%%%%%%%%%%%%%%%%%%% ABSTRACT
\begin{abstract}
Physics informed neural networks (PINNs) have proven to be an efficient tool to represent problems for which measured data are available and for which the dynamics in the data are expected to follow some physical laws. 

In this paper, we suggest a multiobjective perspective on the training of PINNs by treating the data loss and the residual loss as two individual objective functions in a truly biobjective optimization approach. 

As a showcase example, we consider COVID-19 predictions in Germany and built an extended susceptibles-infected-recovered (SIR) model with additionally considered leaky-vaccinated and hospitalized populations (SVIHR model) to model the transition rates and to predict future infections. SIR-type models are expressed by systems of ordinary differential equations (ODEs). 
We investigate the suitability of the generated PINN for COVID-19 predictions and compare the resulting predicted curves with those obtained by applying the method of non-standard finite differences to the system of ODEs and initial data. 

The approach is applicable to various systems of ODEs that define dynamical regimes. 
Those regimes do not need to be SIR-type models, and the corresponding underlying data sets do not have to be associated with COVID-19. 
\end{abstract}

\begin{keyword}
physics-informed neural networks  \sep compartment models
\sep loss function \sep multiobjective optimization \sep weighting parameters \sep Pareto front
\end{keyword}

\end{frontmatter}

%\linenumbers

%%%%%%%%%%%%%%%%%%%%%%%%%%%%%%%%%%%%%%%%%%%%%%%%%%%%%%%%%%%%%%%%%%
\section{Introduction}\label{intro}
Physics informed neural networks (PINNs) \cite{raissi19, blechschmidt2021}, also called theory-inspired machine learning \cite{hoffer22},
have recently become a popular method for solving differential equations. 
By incorporating the residual of the differential equation into the loss function of a neural network-based surrogate model,
PINNs can seamlessly combine measured data with physical constraints given by differential equations. 
PINNs can also be viewed as a \textit{surrogate model} for solving differential equations by incorporating additional data or as a data-driven correction (or even discovery) of the underlying physical system.

\new{By the end of the year 2022, we had experienced several waves of the COVID-19 pandemic with different variants of the virus prevailing at different time intervals. 
Various levels of interventions and protective measures were implemented to counteract the uncontrolled spreading of the disease. 
We focus exemplarily on the time until the fourth wave (i.e., the omicron wave) of the COVID-19 pandemic in Germany that had its peak in February and March 2022.}

The B.1.617.2 (delta) variant of SARS-CoV-2, which is characterized by a higher contagiosity than the previous B.1.1.7 (alpha), B.1.351 (beta) and P.1 (gamma) variants, has been observed
in Germany since March 2021 and was the predominant variant in Germany during several months in the year 2021 \cite{rki3}. 
In the autumn of 2021, the new omicron variant was detected and classified
as concerning by the World Health Organization. 
Three sublines (BA.1, BA.2, BA.3) and a transmission advantage with respect to the delta variant were attributed to the omicron variant at the end of November 2021 \cite{rki3}. 
The omicron variant quickly spread worldwide.
Three recombinations of the omicron and delta variant (XD, XE, XF) had  already been registered as sublines~\cite{rki3}.  

Local peaks during the fourth COVID-19 wave in Germany were reached on November $28^{th}$ 2021 with 693, February $14^{th}$ 2022 with 2,434 and March $20^{th}$ 2022 with 2,619 daily infections per 1 million people. 
All of these peaks were larger than the global peaks of the three previous waves experienced in spring 2020 (69 on April $2^{nd}$ 2020), winter 2020/2021 (305 on December $22^{nd}$ 2020) and spring 2021 (257 on April $25^{th}$ 2021) \cite{ourworldindata}.

The mathematical model used in this work to describe the population dynamics of COVID-19 is the \textit{susceptible-vaccinated-infected-hospitalized-recovered} (SVIHR) model. 
Here, pre-symptomatic individuals are merged with symptomatic people in the infected compartment,
so that we have a single infected compartment of people not hospitalized.  

The contribution of our work consists of \new{two} large parts: 
Firstly, we establish the SVIHR model and build upon this a \textit{Physics-Informed Neural Network} (PINN).
The PINN method uses certain physics-informed constraints, expressed e.g.\ by differential equations, as part of the loss function of a corresponding deep neural network. 
Thus, the system of ODEs plays a crucial role in the training (i.e., the optimization) of the neural network.

PINNs were first introduced in the work of Raissi et al.\ and since then used to solve different forward and inverse problems \cite{raissi19}. 
The PINN approach trades off between the data-based and physical loss functions in the training process. 
This steers the search for reasonable solutions towards those that satisfy a 'physical law' to some degree, i.e., an SVIHR compartmental model in this case. 
The loss function of the PINN is based on reported data of recent infection events (\textit{data loss part}), and a system of ODEs inheriting transition and transmission dynamics, from which so-called residual networks are computed using automatic differentiation (\textit{residual loss part}). The data loss makes PINNs a data-driven technique. Here, we distinguish between training data covering the time since the outbreak of the pandemic in Germany \new{(long-term predictions)} and training data involving exclusively one peak \new{(short-term predictions)}. % so-called long(er)-term and short-term predictions

The PINN involves several fixed model parameters, from which transition parameters are computed. 
\new{A unique feature of our approach is that we estimate the parameters that are crucial to the dynamics of the system using a \emph{nonstandard finite difference} (NSFD) method. 
In other words, we augment the PINN with a numerical approach to the ODE system to better estimate the crucial parameters inside the residual loss function. 
While NSFD solutions generally do not provide good approximations to the data,  they can be used to better estimate model parameters like, e.g., the transmission rates before and after vaccination.}   

The \new{second} contribution of our work is the optimization of the parameter weighting the relation between the data and residual loss part. 
\new{In situations where the physical model can only partially represent the measured data, as is the case when predicting COVID-19 infection rates, data loss and residual loss are in real conflict.
While the main goal is to reproduce the measured data well, the residual loss serves more as a regularization term that helps overcome noise and outliers in the data and better predict the underlying dynamics. 
Choosing a reasonable weighting for these two training objectives is far from trivial.}
To achieve this, we interpret the training process as a biobjective optimization problem, where the residual loss and the data loss are considered as two independent objective functions. 
Rather than combining these two objectives with a pre-determined and fixed weighting parameter, we identify suitable weighting parameters by generating a (rough) approximation of the Pareto front. 

For the training process, we adopt a scalarization-based approach that transforms the biobjective problem into a series of weighted-sum scalarizations. 
Favorable solutions are identified by repeated training runs with adaptively selected scalarization parameters. 
The resulting approximation of the Pareto front provides valuable information on the trade-off between data loss and residual loss. %
On one hand, this information can be used to assess the suitability of the employed physical model. 
On the other hand, a thorough analysis of the (approximated) Pareto front supports an informed selection of a suitable compromise, focusing more on the data or more on the physical model depending on the application background and on the decision makers preferences and beliefs.

%%%%%%%%%%%%%%%%%%%%%%%%%%%%%%%%%%%%%%%%%%%%%%%%%%%%%%%%
\subsection{Related Research}
Since the outbreak of the COVID-19 pandemic, a variety of compartmental models have been introduced as enhanced susceptible-infected-recovered (SIR) compartment models to study various aspects of the spread of SARS-CoV-2. 
PINNs have been applied to compartment models and studied in the context of the COVID-19 pandemic as well.

For instance, Malinzi et al.\ applied a PINN to a susceptible-infected-recovered-deceased (SIRD) model in order to identify the behavioural dynamics of COVID-19 in the Kingdom of Eswatini between March 2020 and September 2021.
They found that their PINN outperformed all other data analysis models even when given minimal quantities of training data \cite{malinzi}.  

Kharazmi et al.\ \cite{kharazmi} considered different integer-order, fractional-order and time-delay models expressed as systems of ODEs.
With the aim of analyzing the past dynamics of COVID-19 in New York City, Rhode Island and Michigan states as well as Italy, they used PINNs that were reported capable of performing parameter inference and simulation of the observed and unobserved dynamics simultaneously. 
Their results showed that purely statistical approaches were generally not well suited for long-term predictions of epidemiological dynamics, and integer-order models seemed to be more robust than fractional-order models, that were first developed by Pang et al.~\cite{pang}. 
Moreover, they stated that no model could accurately capture all the dynamics that play out during an extended pandemic, but models with the ability to adjust key parameters during training could lead to more useful predictions \cite{kharazmi}.   

Cai, Karniadakis and Li calibrated the unknown model parameters of a \textit{susceptible-exposed-infected-removed} (SEIR) model using the novel fractional physics-informed neural networks (fPINNs) deep learning framework in order to obtain reliable short-term predictions of the COVID-19 dynamics caused by the Omicron variant \cite{cai2022fractional}. 
Data from the National Health Commission of the People’s Republic of China covering the time from $27^{th}$ February 2022 to the end of April 2022 were used. For instance, predictions were able to capture sudden changes of the tendency for the new infected cases.

On the other hand, concerning the general PINN approach, 
the multiobjective nature of PINN training was recognized in several recent publications. 
Rohrhofer et al.\ \cite{rohrhofer21} analyze the impact of different weights in a weighted sum objective of data loss and model loss by scanning the weight interval. 
Also Jin et al.\ \cite[Section 4.4.]{Jin2021} studied the 
influence of weights in an experiment for turbulent channel flow,
by manually tuning the weight in order to improve the results. 
Finally, Wang et al.\ \cite[Algorithm~2.1]{wang2021understanding}
proposed an adaptive rule, called 'learning rate annealing for PINNS', for choosing the weights online during the training process.
The basic idea behind this is to automatically tune the weights by using the back-propagated gradient statistics during model training to properly balance all terms in the loss function. 
%}
Their numerical results on diffusion equations and Navier-Stokes equations, respectively, impressively show the impact of the weight selection on the training success. 
Indeed, suboptimal results are obtained for several training runs, thus leaving room for improved multiobjective training approaches.
When the physical model and the data are in good correspondence (this is, for example, the case when the data is artificially generated from the model at hand), an `ideal' solution that simultaneously minimizes data and model loss can be sought. 

Maddu et al.\ \cite{maddu21} suggest a multiobjective descent method that adaptively updates the weights using an inverse Dirichlet strategy to avoid premature termination. While they do not discuss convergence guarantees, their numerical results show a good performance in comparison with recent adaptations of multiobjective descent methods \cite{desideri2012, Fliege2000} to PINN training \cite{sener2019}. 
Stochastic multiobjective gradient descent algorithms were introduced for general NN training in \cite{Liu2019}. 
We also refer to self-adaptive PINNs \cite{McClenny20} and to PINN training in which the loss weights are regarded as hyperparameters \cite{psaros21}.

In a more general setting, multiobjective training approaches were suggested in \cite{reiners} to trade off between data loss and regularization terms in the context of image recognition. 
The different characteristics (slope and curvature) of the considered training goals are addressed by enhancing the stochastic multi-gradient descent approach \cite{Liu2019} with pruning strategies, and by combining adaptive weighted-sum scalarizations with interval bisection. 
The latter supports the identification of favorable knee solutions on the Pareto front.

%%%%%%%%%%%%%%%% Absatz Strukturierung des Papers
This paper is organized as follows: In Section~\ref{sec:modelstructure}, the compartment model for COVID-19 predictions is introduced. 
Firstly, the SIR model is explained in Section~\ref{subsec:sirmodel} to provide an insight into the basics of epidemic modelling. 
Then the system of ODEs of our compartment model, the SVIHR model, is defined along with the used transition rates and transmission rate in Section~\ref{subsec:svihdrmodel}. 
All model parameters are listed in Table~\ref{table:parameters}.

Section~\ref{sec:methods} is devoted to the methodological developments.  
The \textit{Nonstandard Finite Difference} (NSFD) method is introduced in Section~\ref{subsec:nsfd}, where the concept of the scheme is explained, the so-called \textit{denominator function} is derived and the NSFD scheme for the SVIHR model is established. 
Section~\ref{subsec:PINNs} provides an introduction to \textit{physics-informed neural networks} (PINNs) with a focus on the loss function and the suggested neural network structure.
Section~\ref{subsec:paretofront} introduces some aspects of biobjective optimization needed to examine the Pareto front that is obtained by biobjective PINN training approaches. 
Finally, in Section~\ref{subsec:dichotomic}, we introduce a dichotomic search scheme aiming to quickly find near-ideal Pareto optimal solutions and supporting an informed decision on the preferable trade-off between the data loss and the residual loss.

We present our numerical results in Section~\ref{sec:results}. 
\new{In Section~\ref{subsec:valid_short}, we perform a short-term prediction of infection data, using data generated during the delta-variant wave as training data to predict the first omicron wave. 
In Section~\ref{subsec:dichotomicresults}, we continue with the application of the dichotomic search scheme to discuss its ability to approximate a Pareto front, and identify reasonable trade-off solutions. 
Finally, in Section~\ref{subsec:valid_long}, we perform a long-term prediction using most of our available data as training data to predict the delta wave. 
We use the dichotomic search to find a low cost weighting parameter for both objective functions.}
A conclusion is drawn and an outlook to future work is given in Section~\ref{sec:conclusion}.

%%%%%%%%%%%%%%%%%%%%%%%%%%%%%%%%%%%%%%%%%%%%%%%%%%%%%%%%%%%%%%%%%%
\section{A Compartment Model for COVID-19 Predictions}\label{sec:modelstructure} 
The compartment model used to compute the residual loss during PINN training in this paper is the \textit{susceptible-vaccinated-infected-hospitalized-recovered} (SVIHR) model, which was proposed by Treibert and Ehrhardt in \cite{treibert22}. 
It is briefly derived in Section~\ref{subsec:svihdrmodel} again for the sake of completeness.
Building upon the basic \textit{susceptible-infectious-recovered} (SIR) model introduced by Kermack and McKendrick in 1927 \cite{kermack}, the SVIHR model enhances the SIR model to include a vaccinated and a hospitalized compartment. 
A general short introduction to SIR models in mathematical epidemiology is provided in Section~\ref{subsec:sirmodel}. 

In \cite{treibert22}, a comparison between a data-driven PINN approach that takes into account a distinct training data set, and an NSFD method that approximates the SVIHR model was made with regard of the respective prediction qualities for infection and hospitalization numbers. 

Treibert, Brunner and Ehrhardt \cite{treibert21} put the focus on the performance of the NSFD scheme for a \textit{susceptible-vaccinated-infected-intensive care-deceased-recovered} (SVICDR) model. 
Here, the impact of modifications parameter bounds on the predicted prevalence was investigated, taking into account data from the pandemic in Germany and an exponentially increasing vaccination rate in the considered time window as well as trigonometric contact and quarantine rate functions.
The results showed that the NSFD methods can predict a global peak solely based on the mathematical model and the defined parameters, but independently of a previously observed behavior of the infectious disease.

\new{In this paper, we build on the SVIHR model of \cite{treibert22}.
 A novel PINN approach is presented based on updated data from \cite{rki1, rki2, impfdashboard} that incorporates both short-term and long-term data for the predictions. 
An improved network architecture is complemented by a dynamics-based parameter estimation that combines NSFD and PINN methods.
 Characteristic for the considered application is the often significant deviation of the measured data from the model predictions: 
While the SVIHR model captures the dynamics of COVID-19 infection, it cannot reflect fluctuations in the data that may have a variety of different causes. 
We address this challenge with an adaptive approach to analyze the trade-off between data loss and residual loss in the training process.
 This biobjective perspective on PINN training enables semi-automatic and problem-specific identification of optimized weighting parameters.} 

As determining the proportion of asymptomatic individuals in the total infected population is not our goal at this point, 
we do not incorporate a separate compartment of asymptomatic infected individuals, 
but assume at least very mild symptoms in infected individuals. 
The degree of infectivity of infected individuals can be regulated by adjusting the transmission rate in the model. 
Our model is adaptable to different vaccination and transmission scenarios. 

%%%%%%%%%%%%%%%%%%%%%%%%%%%%%%%%%%%%%%%%%%%%%%%%%
\subsection{The SIR Model in Mathematical Epidemiology}\label{subsec:sirmodel} 
The basic SIR model consists of three compartments of susceptible ($S$), infected ($I$), and recovered ($R$) individuals.
We denote with $K(t)$ the size of a compartment $K\in\{S,I,R\}$ at time $t$, where a time unit equals a week.
Susceptible individuals have not yet become infected but may become ill. 
In the basic SIR model, infected individuals may infect susceptible persons, 
i.e.\ they are assumed to be infectious (without any delay)  
and may or may not have symptoms. 
Recovered individuals have overcome the disease and are assumed to be neither infectious nor ill. 

The total size of the population at time $t$ is denoted by $N(t)$. 
The satisfaction of the equation
\begin{equation*}
    N(t) = S(t)+I(t)+R(t) \quad \text{ with } N\colon[0,T]\to\mathbb{N},
\end{equation*}
means that the number of individuals in the system is the sum of the compartment sizes at each considered time point  $t\in[0,T]$.
The system \eqref{eq:table0} must have initial conditions $S_0=S(0)$, $I_0=I(0)$, $R_0=R(0)=0$ to be well-defined \cite[p. 11]{maia}.  
The population size $N(t)$ is assumed to be constant, i.e.\ $N(t)=N$, and the derivative of $N(t)$ is zero, which means %in case 
that the system does not consider a recruitment rate $\Lambda$ nor a natural death rate.

Let $p$ be the probability that a contact with a susceptible individual results in a transmission, and let $\zeta$ be the per capita contact rate, i.e.\ the number of contacts made by one infectious individual. 
Then $\zeta\,N$ is the number of contacts per unit of time this infectious individual makes, and $\zeta\, N\,\frac{S}{N}$ denotes the number of contacts with susceptible individuals that one infectious individual makes per unit of time.
Moreover, we define a transmission rate constant $\beta$ \cite[p. 10]{maia} as  
\begin{equation}
    \beta = p \, \zeta \, .
\end{equation}
\new{For a more detailed discussion of the transmission rate and of related parameters including, among others, time dependent models, we refer to \cite{treibert21}.}

If $I(t)$ stands for the number of infected individuals at time $t$ (\textit{prevalence)}, then $\beta\,S\,I$ denotes the number of individuals who become infected per unit of time (\textit{incidence}). 
If $\omega_I$ is the recovery rate, we obtain the following system of ODEs, that describes the SIR model \cite[p. 11]{maia}:  
\begin{equation}\label{eq:table0} 
    \begin{split}
    \dt{S(t)} &= -\beta \, I(t) \, S(t),\\
    \dt{I(t)} &= \beta \, I(t) \, S(t) - \omega_I \, I(t), \\ 
    \dt{R(t)} &= \omega_I \, I(t). 
 \end{split}
\end{equation} 

For the model in Equation \eqref{eq:table0}, the maximum number of infected individuals that can be reached in the regarded epidemic is bounded by 
\begin{equation}
    I_{\max} = - \frac{\omega_I}{\beta} + \frac{\omega_I}{\beta} \, \ln\bigl(\frac{\omega_I}{\beta}\bigr) + S_0 + I_0 - \frac{\omega_I}{\beta} \, \ln\bigl(S_0 \bigr) \, .
\end{equation} 
Let 
\begin{equation} 
      F(t)=1-e^{-\omega_I t}, \quad t \ge 0 \, 
\end{equation} 
be the probability of recovering/leaving the infectious compartment in the time interval $[0,t)$ \cite[p. 11]{maia}. 
The function $F(t)$, with $F(t)=0$ for $t<0$, is a probability distribution. 
Then $f(t)=\dt{F(t)}$ is the respective probability density function: 
\begin{equation}\label{de}
   f(t)=\omega_I \, e^{-\omega_I t}  \new{\quad \text{for}\quad t> 0,\;\text{and} \quad f(t)=0 \quad \text{for}\quad t\le0.}
\end{equation}
If $X$ denotes the average time spent in the infectious compartment, then the mean time spent in the infectious compartment can be computed as the first moment
\begin{equation}
   \mathbb{E}\bigl[X\bigr] = \int_{-\infty}^\infty t \, f(t)\, dt 
   = \int_{\new{0}}^\infty t \, \omega_I \, e^{-\omega_I t}\, dt = \frac{1}{\omega_I} \, .
\end{equation} 
For SARS-CoV-2, the mean time of infectiousness is not clearly defined. 
With a mild or moderate course of the disease, contagiousness clearly declines within the ten days after symptom occurrence. 
Contagiousness has to be distinguished from positive test results, that can occur several weeks after catching the infection, although the infectiousness is usually on a very low level then \cite{episteckbrief}.

In this basic form of the SIR model, the population is assumed to be closed so that no individual enters or leaves a compartment from the outside, and recovered individuals are completely immune so that they can never be reinfected \cite[p. 13]{maia}. 

%%%%%%%%%%%%%%%%%%%%%%%%%%%%%%%%%%%%%
\subsection{The SVIHR Model}\label{subsec:svihdrmodel}
The SIR model was enhanced by a compartment of hospitalized individuals $H$ and a compartment of fully, 
i.e.\ at least twice, vaccinated individuals $V$ in \cite{treibert22}. 
Data was obtained from the Robert Koch-Institute (RKI) \cite{rki1, rki2} and the German COVID-19 Vaccination Dashboard \cite{impfdashboard}, based on which parameter values and compartment sizes, referred to as \textit{reported compartment sizes} or \textit{reported data} in the sequel, were computed.

Infected individuals remain infected for $T_I$ days until they recover, where a proportion $\xi$ of all  
individuals transiting from the infected individuals are hospitalized. 
The exact daily or weekly number of infectious individuals among the infected individuals is not known.
The number of infections registered by the Robert Koch-Institute (RKI) is used to compute the reported size of the compartment $I$ for all considered calendar weeks in this paper. 
This number is based on the number of infected individuals who are infectious enough so that the virus is usually verifiable via a rapid antigen test. 
Infectious and not infectious infected people are merged within the compartment $I$. 
The general degree of infectiousness of the individuals in $I$ depends on the transmissibility of the virus and is included in the transmission rate. 

According to the RKI, the concrete time period of contagiosity is not clearly defined, but infectiousness is highest right before and after the presence of first symptoms and drastically declines after at most $10$ days after the very first symptoms occur (assuming a mild or moderate course of disease) \cite{episteckbrief}. 
We set $T_I=1.2$ weeks, i.e.\ 8.4 days, to adopt for a small time span of 1-2 days between the first showing of symptoms and getting tested. 
The parameter $\omega_1$ is the rate at which individuals per unit time (week) pass from compartment $I$ to $R$. 
It is defined as 
\begin{equation}
    \omega_1 = \frac{1-\xi}{T_I} \, .        
\end{equation}  
The rate $\eta$ at which individuals reach the compartment $H$ per unit of time is defined as
\begin{equation}
    \eta = \frac{\xi}{T_I}.
\end{equation} 
%As a proportion $\xi$ of currently infected individuals is assumed to be hospitalized within $T_I$ weeks, a proportion $1-\xi$ is assumed to recover within those $T_I$ days, such that 
Thus, we assume that $\omega_1 \, I(t)$ people recover and $\eta \, I(t)$ individuals are hospitalized within week $t$. 
It is assumed here that hospitalized individuals cannot infect susceptible individuals because they are isolated.  
They are assumed to remain infected $T_H$ days from the time of their hospitalization. 
In a German academic survey with 1,426 COVID-19 patients with an acute respiratory disease, an average duration of hospital stay of 10 days was observed %the result
\cite{episteckbrief}. Accordingly, we set $T_H=1.5$ weeks, i.e.\ 10.5 days. 

The vaccinated compartment $V$ contains all susceptible individuals who have received a COVID-19 vaccination. 
It is reached from the compartment $S$ at a rate $\mathcal{V}$. 
If vaccination does not guarantee complete immunity to infection, we speak of a \textit{leaky vaccination}. 
Due to the assumed leakiness, all vaccinated individuals have a lower probability of
contracting the infection than susceptible individuals in compartment $S$.  
If an all-or-nothing vaccine was assumed, vaccinated people would be completely protected
from the infection to a specific portion of the susceptible class per unit time $t$,
whereas the other susceptible individuals did not gain any protection. 
Let $\kappa$ denote the residual probability of infection after vaccination. 
The rate at which vaccinated individuals reach the infected compartment $I$ is then \new{$\kappa\beta I(t)$.} %$\kappa\,\theta(t)$.

Furthermore, we incorporate a constant system inflow, the so-called recruitment rate $\Lambda$ (e.g.\ birth of new individuals that can get infected), and the natural mortality rate $\mu$. 
The recruitment and natural death rate are set to zero as they are regarded as equal in both \cite{treibert22} and in this paper, but are still included in the system of ODEs for the purpose of properly deriving the denominator function in the NSFD scheme, see Section~\ref{subsec:nsfd}. 
The total population size is kept constant like this.   
The corresponding system of ODEs has the following form: 
\begin{equation}\label{eq:table1} 
    \begin{split}
    \dt{S(t)}&= \Lambda 
    %- \beta \, \zeta \, \bigl(1-q \bigr) \, \frac{I(t)S(t)}{N(t)} 
    \new{- \beta I(t)S(t)}
    - (\mathcal{V}+\mu) \, S(t),\\
  \dt{V(t)} &= \mathcal{V} \, S(t) 
  %- \beta \, \zeta \, \bigl(1-q \bigr) \, \kappa \, \frac{I(t)S(t)}{N(t)}
  \new{-\kappa \beta I(t)S(t)}
  -\mu\, V(t),\\ 
    \dt{I(t)} &= %\beta \, \zeta \, \bigl(1-q \bigr)  \, \bigl(1+\kappa\bigr) \frac{I(t)S(t)}{N(t)} 
    \new{(1+\kappa)\beta I(t)S(t)}
    - \bigl(\eta + \omega_1 +\mu\bigr) \, I(t), \\
    \dt{H(t)} &= \eta \, I(t) -  (\omega_2+\mu) \, H(t),\\
    \dt{R(t)} &= \omega_1 \, I(t) + \omega_2 \, H(t)-\mu \,R(t). 
 \end{split}
\end{equation}  
The system~\eqref{eq:table1} extends the simple system~\eqref{eq:table0} by the differential equations
for $V(t)$ and $H(t)$, 
describing the inflow into and the outflow from the compartment $V$ or $H$, respectively, 
as well as the recruitment rate $\Lambda$ and the natural death rate $\mu$. 
\new{Note that the equation for $V(t)$ in \eqref{eq:table1} can only approximate the actual dynamics, since vaccinated individuals remain vaccinated even when they become infected. 
We approximate the respective rates of change by using the compartment of susceptible individuals as a reference in the term $\kappa\beta I(t)S(t)$. 
Preliminary numerical tests, using both the NSFD scheme and the PINN approach, show that the use of $\kappa\beta I(t)V(t)$ does not significantly affect the quality of the predictions.}

\new{One can easily show that for positive parameters and positive initial data, the solution of \eqref{eq:table1} remains positive for all times.
We call this the "positivity property".
If we now set $\Lambda=\mu=0$ and add all compartments in \eqref{eq:table1}, we find that the total population is a quantity conserved over time. 
We will return to this "conservation property" later when we discuss the NSFD scheme.
This setting is an acceptable simplification because the time scale of human births and deaths is much longer than that of a COVID-19 epidemic wave.}

\begin{figure}[h!]  
\begin{center}
\begin{tikzpicture}
\node (A) at (2,-4) [circle,shade,draw] {S};
\node (I) at (4,-4) [circle,shade,draw] {I};
\node(C) at (8,-4) [circle,shade,draw] {H};
\node (K) at (6,-3) [circle,shade,draw] {R};
\node (V) at (2,-5.5) [circle,shade,draw] {V};  
\draw[->, blue!50, very thick] (A) to
node[left] {$\mathcal{V}$} (V); 
\draw[->, blue!50, very thick] (A) to node[below] {$\beta I(t)$} (I); 
\draw[->, blue!50, very thick] (V) to node[right] {$\kappa \, \beta I(t)$} (I);
\draw[->, blue!50, very thick] (I) to[bend left=20] node[below] {$\omega_1$} (K); 
\draw[->, blue!50, very thick] (C) to[bend right=20] node[below] {$\omega_2$} (K); 
\draw[->, blue!50, very thick] (I) to node[below] {$\eta$} (C); 
\draw[->, blue!50, thick] (I) to[bend right=50](A);
\draw[->, blue!50, thick] (A) to node[right] {$\mu$} (2,-2.75);
\draw[->, blue!50, thick] (I) to node[right] {$\mu$} (4,-2.75);
\draw[->, blue!50, thick] (C) to node[right] {$\mu$} (8,-2.75);
\draw[->, blue!50, thick] (K) to node[right] {$\mu$} (6,-1.75);
\draw[->, blue!50, thick] (V) to node[right] {$\mu$} (2,-6.75); 
\draw[->, blue!50, thick] (0.5,-4) to node[above] {$\Lambda$} (A);
\end{tikzpicture} 
\caption{Illustration of the compartments and their interrelation in the SVIHR model} 
\label{fig:modeldynamics}    
\end{center}
\end{figure}
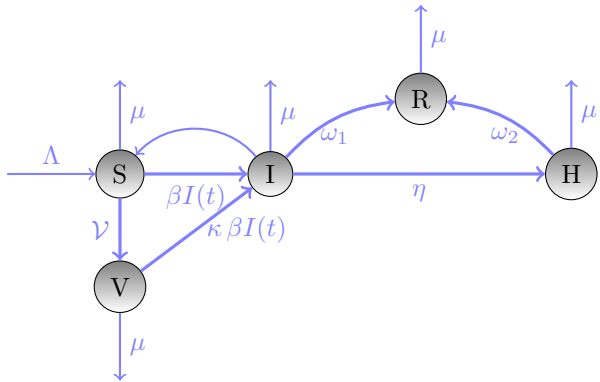

Figure~\ref{fig:modeldynamics} shows the dynamical system described by %equation 
\eqref{eq:table1}.
Blue arrows from one compartment to another indicate a transition.
Table~\ref{table:parameters} lists the model parameter definitions and used values. We note that the parameter values for the SVIHR model used in this paper stated in Table~\ref{table:parameters} differ from the ones used in \cite{treibert22}.

\begin{sidewaystable}[htp]
\centering
%\begin{table}
\small 
% \footnotesize %\hspace{-3cm}
\begin{tabular}{cccc}
    \textbf{Parameter} & \textbf{Definition} & \textbf{Parameter Value used in \ref{subsec:valid_long}} & \textbf{Parameter Value used in \ref{subsec:valid_short} and \ref{subsec:dichotomicresults}} \\
     $N$ & total population size & 83,100,000 \cite{destatis} & 83,100,000 \cite{destatis} \\
     $\beta$ & transmission risk parameter & \textit{determined via \new{NSFD}} & \textit{determined via \new{NSFD}} \\
     $\kappa$ & residual infection prob.\ after vaccination & \textit{determined via \new{NSFD}} & \textit{determined via \new{NSFD}} \\ 
     $T_I$ & average length of infection period & 1.2 \cite{episteckbrief} &  1.2 \cite{episteckbrief} \\
     $T_H$ & average length of hospitalization period & 1.5 \cite{episteckbrief} & 1.5 \cite{episteckbrief} \\
     $\mathcal{V}$ & vaccination coefficient & 0.0159 \cite{impfdashboard} & 0.0231 \cite{impfdashboard} \\
     $\xi$ & hospitalization coefficient & 0.0862 \cite{rki2} & 0.0735 \cite{rki2} \\
     $\mathcal{M}$ & mortality coefficient & 0.0232 \cite{rki1} & 0.0142 \cite{rki1} \\  
     $\omega_1$ & weekly recovery rate for infected people & 0.7615 & 0.7721 \\
     $\omega_2$ & weekly recovery rate for hospitalized people & 0.6512 & 0.6572 \\  
     $\eta$ & weekly hospitalization rate & 0.0719 & 0.6125 \\
     % $\mu_1$ & disease-induced mortality rate & 0.0155 \\  
\end{tabular}
\caption{Parameters of the SVIHR model and their values in the implementation of the PINN.}
\label{table:parameters}
%\end{table}
\end{sidewaystable}

%%%%%%%%%%%%%%%%%%%%%%%%%%%%%%%%%%%%%%%%%%%%%%%%%%%%%%%%%%%%%%%%%%%
\section{Finding Optimized Weights in PINN Approaches}\label{sec:methods} 
We implement a physics informed neural network (PINN) that is trained both w.r.t.\ German COVID-19 data and w.r.t.\ the SVHIR model introduced in Section~\ref{subsec:svihdrmodel} above. 
This technique is validated using error computations with regard to reported data.
The method and structure of PINNs is explained in Section~\ref{subsec:PINNs}.
Moreover, scenarios generated using the PINN are compared to those produced using the technique of nonstandard finite difference (NSFD) schemes. 
A short introduction to NSFD schemes and the application on the SVIHR model are outlined in Section~\ref{subsec:nsfd}.

Since the PINN is trained w.r.t.\ two loss terms, we take a biobjective approach to investigate the influence of weighting parameters on each loss term.
We consider the two losses as independent objective functions and want to find weighting parameters to achieve an approximation of the Pareto front. 
We therefore first introduce certain aspects of bicriteria optimization in Section~\ref{subsec:paretofront} and then introduce a dichotomic search scheme to efficiently approximate the Pareto front in Section~\ref{subsec:dichotomic}.

%%%%%%%%%%%%%%%%%%%%%%%%%%%%%%%%%%%%%%%%%%%%%%%%%%%%%%%%%%
\subsection{Nonstandard Finite Difference Schemes}\label{subsec:nsfd}
NSFD schemes go back to a paper by Mickens published in 1989 \cite{mickens1}. 
Their structural properties originate from investigations of special groups of differential equations for which exact finite difference schemes are not available. 
For the sake of completeness, the NSFD scheme for the SVIHR model is described again here. 
It has already been derived in a similar way in \cite{treibert22}.  

In NSFD schemes, derivatives have to be modelled by proper discrete analogues,
i.e.\ nonstandard difference quotients
of the form, cf.\ \cite{mickens2}
\begin{equation}\label{eq:discrete}
    \frac{\text{d}u(t)}{\text{d}t} \to \frac{u_{n+1}-\psi(h)u_n}{\phi(h)} \, ,
\end{equation} 
where $t_n=n\,h$, $u_n$ is the approximation of $u(t_n)$, and $\psi(h) = 1 + \mathcal{O}(h)$. 
\new{$\phi(h)$ is a denominator function, which is explained in more detail below.}
Using this rather general time discretization \eqref{eq:discrete} in NSFD schemes our aim is to model the asymptotic long-time behaviour of the solution. 
A numerical scheme for a system of first-order differential equations is called NSFD scheme if at least one of the following conditions hold \cite{mickens2}:
\begin{itemize}
\item Discrete representations for derivatives must, in general, have nontrivial denominator functions. 
Here, the first-order derivatives in the system are approximated by the generalized forward difference method %(forward Euler method) 
$\frac{\text{d}u_n}{\text{d}t} \approx \frac{u_{n+1}-u_n}{\phi(h)}$, where $u_n\approx u(t_n)$ and $\phi \equiv \phi(h)>0$ is the so-called \textit{denominator function} such that $\phi(h)=h+\mathcal{O}(h^2)$, with $h$ the step size.  

\item The consistency orders of the finite difference quotients % discrete derivatives 
should be equal to the orders of the corresponding derivatives appearing in the differential equations.
\item The nonlinear terms are approximated by non-local discrete representations, for instance by a suitable function of several points of a mesh, 
like $u^2(t_n) \approx u_n u_{n+1}$ or $u^3(t_n) \approx u^2_n u^{\phantom{2}}_{n+1}$.
\item Special conditions that hold for either the ODE and/or its solutions should also hold for the difference equation model and/or its solution, e.g.\ the equilibrium points of the underlying ODE system.
\end{itemize}

%%%%%%% Matthias
\new{In contrast to conventional difference methods, NSFD schemes focus not only on stability and convergence order, but also on qualitative properties, i.e., how well the discrete model (the NSFD scheme) reproduces the most important properties of the underlying continuous model.}
NSFD schemes preserve the positivity property and satisfy the conservation law for $\Lambda=\mu=0$ yielding the stability of the scheme. 
The equilibrium points of the ODE model \eqref{eq:table1} appear in the proposed NSFD-scheme as well.  

In the sequel, we will derive an appropriate \textit{denominator function} $\phi(h)$ for the NSFD discretization of the system~\eqref{eq:table1}. This function is chosen such that the numerical solution exhibits the same asymptotic behaviour as the analytic solution. 
To do so, we
consider the total population $N=S+V+I+H+R$ of the ODE system \eqref{eq:table1}. 
\begin{comment}
We do not neglect the recruitment rate $\Lambda$ and natural death rate $\mu$ here. 
Thus, we obtain the SVIHR model described by the system of ODEs 
\begin{equation}\label{eq:svihr} 
    \begin{split}
    \frac{\text{d}S(t)}{\text{d}t} &= \Lambda- \theta(t) \, \frac{S(t)}{N(t)} - (\mathcal{V}+\mu) \, S(t),\\
    \frac{\text{d}V(t)}{\text{d}t} &= \mathcal{V} \, S(t) - \theta(t) \, \kappa \, \frac{S(t)}{N(t)}-\mu\, V(t),\\ 
    \frac{\text{d}I(t)}{\text{d}t} &= \theta(t) \, \bigl(1+\kappa\bigr) \frac{S(t)}{N(t)} - \bigl(\eta + \omega_1 +\mu\bigr) \, I(t), \\
    \frac{\text{d}H(t)}{\text{d}t} &= \eta \, I(t) -  (\omega_2+\mu) \, H(t),\\
    \frac{\text{d}R(t)}{\text{d}t} &= \omega_1 \, I(t) + \omega_2 \, H(t)-\mu \,R(t) \, .  
 \end{split}
\end{equation}  
\end{comment}
%
Adding the equations of \eqref{eq:table1}, a differential equation
describing the dynamics of the total population $N$ is obtained as 
\begin{equation}\label{nsfd_1}
   \frac{\text{d}N(t)}{\text{d}t} = \Lambda-\mu\,N(t) \, , 
\end{equation}
which is solved by 
\begin{equation}\label{nsfd_2}
    N(t) = \frac{\Lambda}{\mu} + \Bigl(N(0)-\frac{\Lambda}{\mu}\Bigr) \,e^{-\mu t} = N(0) +
    \Bigl(N(0)-\frac{\Lambda}{\mu}\Bigr) \,(e^{-\mu t}-1),
\end{equation}
where $N(0)=S(0)+V(0)+I(0)+H(0)+R(0)$.
\new{Similar to \eqref{nsfd_1}, adding the equations of an NSFD discretization of \eqref{eq:table1} yields the equation} 
\begin{equation}\label{nsfd_4} 
   \frac{N^{n+1}-N^n}{\phi(h)} = \Lambda - \mu\,N^{n+1}, 
\end{equation} i.e.
\begin{equation}\label{nsfd_4b} 
\begin{split}
   N^{n+1}&= \frac{N^n +\phi(h)\Lambda}{1+\phi(h)\,\mu}
   =N^n-\Bigl(N^n -\frac{\Lambda}{\mu}\Bigr)
   \frac{\phi(h)\,\mu}{1+\phi(h)\,\mu}\\
   &=N^n+\Bigl(N^n -\frac{\Lambda}{\mu}\Bigr)
   \Bigl(\frac{1}{1+\phi(h)\,\mu}-1\Bigr).
   \end{split}
\end{equation} 
The denominator function can be derived by comparing Equation \eqref{nsfd_4} with the discretized version of Equation \eqref{nsfd_2}, that is
\begin{equation}\label{nsfd_3}
    N^{n+1} =  N^n + \Bigl(N^{n}-\frac{\Lambda}{\mu}\Bigr) \,(e^{-\mu h} - 1), 
    \quad h=\Delta t,
\end{equation} 
such that the (positive) denominator function is defined by 
\begin{equation}
    \frac{1}{1+\phi(h)\,\mu}=e^{-\mu t},
\end{equation}
i.e.\
\begin{equation}\label{denom}
\phi(h) = \frac{e^{\mu h}-1}{\mu} 
= \frac{1+\mu h + \frac{1}{2} \mu^2  h^2  + \ldots - 1}{\mu} = h + \frac{\mu h^2}{2} + \ldots= h + \mathcal{O}(h^2). 
\end{equation}       

Now, making use of the denominator function in \eqref{eq:discrete}, the NSFD discretization can be established. 
The implicit form of this discretization is provided in Equation \eqref{eq:table3}. Here, $\phi(h)$ is given by \eqref{denom} 
\begin{flalign}\begin{split}\label{eq:table3} 
   \frac{S^{n+1}-S^n}{\phi(h)} &= \Lambda-\beta\, I^n \,S^{n+1} - (\mathcal{V} + \mu) \,S^{n+1},\\
\frac{V^{n+1}-V^n}{\phi(h)} &=  \mathcal{V} \,S^{n+1} - \beta \,\kappa \, I^n \,S^{n+1} - \mu \, V^{n+1},\\
\frac{I^{n+1}-I^n}{\phi(h)} &=  \beta \,(1+\kappa) \, I^{n+1} \, S^{n+1} - (\eta+\omega_1+\mu) \,I^{n+1},\\
\frac{H^{n+1}-H^n}{\phi(h)} &=  \eta\, I^{n+1} - (\omega_2+\mu)\, H^{n+1},\\
%\frac{H^{n+1}-H^n}{\phi(h)} &=  \eta\, I^{n+1} - (\omega_2+\lambda+\mu)\, H^{n+1},\\
%\frac{D^{n+1}-D^n}{\phi(h)} &=  \lambda \, H^{n+1},\\
\frac{R^{n+1}-R^n}{\phi(h)} &=  \omega_1 \,I^{n+1} + \omega_2 \, H^{n+1} - \mu \, R^{n+1} \, .
\end{split}\end{flalign} 
We can rewrite the scheme in order to obtain an explicit variant of it, as to be found in \eqref{eq:table2}. 
From the explicit representation we can deduce that this scheme preserves the positivity.
\begin{flalign}\begin{split}\label{eq:table2} 
            S^{n+1} &= \frac{S^n +\phi(h)\Lambda}{1+\phi(h) \,(\beta\, I^n+\mathcal{V}+\mu)}, \\
            V^{n+1} &= \frac{V^n+\phi(h) \, S^{n+1} \, (\mathcal{V}-\beta \, \kappa \, I^n)}{1+\phi(h) \, \mu}, \\
             I^{n+1} &= \frac{I^n}{1+\phi\, (\eta+\omega_1+\mu - \beta \,(1+\kappa) \, S^{n+1})},\\
            H^{n+1} &= \frac{\phi(h) \,\eta\, I^{n+1}+H^n}{1+\phi\,(\omega_2+\mu)}, \\
            R^{n+1} &= \frac{R^n+\phi(h) \,(\omega_1\, I^{n+1}+\omega_2\, H^{n+1})}{1+\phi(h) \,\mu} \, .
\end{split}\end{flalign}   
The calculation must be implemented in exactly this order. 
All parameters appearing in these type of epidemic models are always non-negative. 

\new{
 \begin{rmk}
 Looking at the third equation of \eqref{eq:table3}, one might wonder why one does not use the time discretization $I^n S^{n+1}$ instead of $I^{n+1}S^{n+1}$.
 Here two arguments meet in this discussion, and there is no clear right or wrong. 
 With this "new" version, one has the exact conservation property on the discrete level. 
 With the "old" version $I^{n+1}S^{n+1}$ (implicitly only in $I^{n+1}$, $S^{n+1}$ is already calculated in the first line), on the other hand, one gets positivity conservation and thus also the stability property, which is more important. 
So we have chosen the first variant and have to cope with a small perturbation of the conservation of the total population.
\end{rmk}
}
\new{For an actual application, where a standard solver gives false negative solution and this problem can be solved by a NSFD scheme
we refer to \cite{maamar2022}.}

%%%%%%%%%%%%%%%%%%%%%%%%%%%%%%%%%%%%%%%%%%%
\subsection{Physics-informed Neural Networks for Compartment Models}\label{subsec:PINNs} 
\textit{Physics-informed neural networks} (PINN) are neural networks that include the laws of dynamical systems into a deep learning framework. 
Machine learning has emerged as an alternative to numerical discretization in high-dimensional problems governed by partial differential equations. 
Nonetheless, a sufficient amount of data as required for training deep neural networks is not necessarily available. 
In such cases, missing data can be substituted by incorporating additional information obtained from enforcing the physical laws of dynamical systems \cite{karniadakis1}. Such laws can be described by partial or ordinary differential equations.
One example where  dynamical systems can be used are populations undergoing transitions between different infected or uninfected states during an epidemic, as considered in this paper. 

PINNs can approximate the solutions of differential equations by training a loss function incorporating the initial and boundary conditions and the residual at so-called collocation points \cite{cuomocola}. 
Instead of approximating solutions of differential equations, PINNs can use a system of differential equations describing a certain real-world process along with time-series data sets describing the past course of such a process for the purpose of %drawing scenarios 
generating predicitions for  future progressions. 

The loss function of a corresponding neural network includes not solely the so-called \textit{data error} related to the difference between the output of the network and the reported data used, but also the so-called \textit{residual error} related to the ODEs or PDEs. 

Olumoyin et al.~\cite{olumoyin21} refer to a type of feedforward neural network including epidemiological dynamics such as lockdown into their loss function by using the term \textit{Epidemiology-Informed Neural Network (EINN)}. 
EINNs extend PINNs for epidemiology models and are able to capture the dynamics of the spread of the disease and
the influence of the mitigation measure. 
The loss function is enhanced to include time-varying rates using epidemiology facts about the infectious disease \cite{olumoyin21}. 

Shaier, Raissi and Seshaiyer~\cite{shaier21} describe a type of PINN-based neural network that can be applied to increasingly complex systems of differential equations describing various known infectious diseases with the term \textit{Disease-Informed Neural Networks (DINN)}. 
DINNs can be systematically applied to increasingly complex governing systems of differential equations describing infectious diseases. 
They are able to effectively learn the dynamics of the disease and forecast its progression a month into the future from real-life data \cite{shaier21}. 

The neural network established and applied in this paper can be described as a special type of EINN or DINN based on a special kind of epidemic compartment model called SVIHR model regarding the susceptible, vaccinated, infected, hospitalized and recovered part of the population. We focus on the German population in this paper, using data provided by the Robert Koch-Institute \cite{rki1, rki2} and the German COVID-19 Vaccination Dashboard \cite{impfdashboard}. The model parameters included in \eqref{eq:table1} can be comprised in a vector $\vartheta$: 
\begin{equation}
    \vartheta=[\beta, \mathcal{V}, \kappa, \xi, T_I, T_H, \mathcal{M}, T_H]^\top\,. 
\end{equation}
\new{This vector is partitioned into fixed parameters $p_f$, which can be estimated directly from the observed data, and learnable parameters $p$, which are crucial for the dynamics of the compartment variables:}
\begin{equation} \label{eq:p}
\begin{split}
     p_f &:= [\mathcal{V}, \xi, T_I, T_H, \mathcal{M}, T_H]^\top , \\ 
     p &:= [\beta, \kappa]^\top\,.
\end{split} 
\end{equation}
\new{We use NSFD predictions to improve the estimates for the parameters that govern the dynamics, i.e.,} the transmission risk $\beta$ and the residual transmission probability after vaccination $\kappa$. 
\new{Their values are estimated using NSFD predictions fitted to the observed data so that the predicted and observed peak values have the same magnitude. 
This approach aims to combine the benefits from both worlds: 
The numerical predictions that follow a physical model, and the PINN predictions that rely heavily on the observed data. 
Since the NSFD predictions can be computed very efficiently, the parameter values can be easily adapted to different virus variants and to different pandemic waves.}

\new{We describe our PINN approach for a general compartmental model with $n$ compartments. 
In the SVIHR model \eqref{eq:table1}, we have $n=5$. Let}
\begin{equation*}
    \hat{\mathcal{K}}(t)=[\hat{\mathcal{K}}^1(t),\dots,\hat{\mathcal{K}}^n(t)]^\top
\end{equation*}
\new{be the normalized data vector of reported compartment sizes for $l$ time points, i.e.\ $t\in\{t_1,\dots,t_l\}$. 
In the application considered here, $t_{i+1}-t_i$ is constant (and corresponds to one week) for all $i=1,\dots,l-1$.  
Let $T$ denote an upper bound for the time window under consideration, and} 
let $\mathcal{K}_p:[0,T]\to\R^n$ denote the vector valued function reflecting the compartment sizes $\mathcal{K}_p(t)$ over time. 
Then we express the vector of right-hand-sides values of \eqref{eq:table1} by 
\begin{equation*}
    F(\mathcal{K}_p)=[F^1(\mathcal{K}_p),\dots,F^n(\mathcal{K}_p)]^\top \,.
\end{equation*} 
The subscript $p$ stands for the learnable model parameters that the system of ODEs depends on, cf.\ \eqref{eq:p}, and thus the solution will depend on, too.
The system of ODEs in \eqref{eq:table1} can then be discretized as 
\begin{equation}
    \dt{\mathcal{K}_p(t)}-F(\mathcal{K}_p)=0, \qquad t \in \{t_1,\dots,t_l\} \, .
\end{equation}
Our PINN 
\begin{equation*}
    \mathcal{PINN}_p^{W}\colon\new{[0,T]} \to\mathbb{R}^n
\end{equation*} 
is used to approximate the solution $\mathcal{K}_p:[0,1]\rightarrow\R^n$ 
of the system of ODEs \eqref{eq:table1} by performing error minimization during training \cite{grimm}. 
The superscript $W$ represents the weights used during the forward and backward propagation in the neural network. 
At time instance $t$, $t\in\{t_1,\dots,t_l\}$, the solution is expressed as
\begin{equation*}
    \mathcal{K}_p(t)=[\mathcal{K}_p^1(t),\dots,\mathcal{K}_p^n(t)]^\top \, ,
\end{equation*}
where $\mathcal{K}_p^j(t)$ is the output of the PINN for the $j^{th}$ compartment at time $t$ 
\new{and depends on the parameter $p$ given by equation \eqref{eq:p}}.  
The parameters $W$ %and $p$ %$p_t(t)$ 
are optimized during the backpropagation process of the neural network such that $\PINNpW$ fits the reported data $\hat{\mathcal{K}}$ in a least-squares sense \cite{grimm}. 
In the \new{$i^{th}$ time step $t_i$, $i \in \{1,\dots,l\}$},
with PINN output $\PINNpW(\new{t_i})$, we compute the usual \textit{data error} defined as 
\begin{equation}\label{eq:dataerror}
  \MSEU =\MSEU(W) := \frac{1}{\new{l}}\sum_{\new{i}=1}^\new{l} \|\PINNpW(\new{t_i})-\hat{\mathcal{K}}_\new{p}(\new{t_i})\|^2\,,
\end{equation}
\new{where we employed in \eqref{eq:dataerror} the Euclidian norm in $\R^n$.}

Next, let us 
extend the loss function of the PINN by the additional term 
\begin{equation}\label{eq:residualerror}
    \mathcal{F}_p(\PINNpW,\new{t_i})
    :=\dt{\PINNpW(\new{t})}\Big|_{t=\new{t_i}} - F_p\bigr(\PINNpW(\new{t_i})\bigr)\, ,
\end{equation} 
where 
\begin{equation}
    \mathcal{F}_p(\mathcal{PINN}_p^W,\new{t_i}) = 0 \quad\text{for all} %\forall 
    \; t \in \{t_1,\dots,t_l\}
\end{equation}
\new{means that the PINN solves the given ODE system more reliably by enforcing additional constraints.}
The computation of the time derivative of the neural network output $\dt{\PINNpW(\new{t})}\Big|_{t=\new{t_i}}$ can be performed using automatic differentiation \cite{AD}. 
Then the physics-informed part of the loss function, the \textit{residual error}, is given by 
\begin{equation}
  \MSEF =\MSEF(W) := \frac{1}{\new{l}}\sum_{\new{i}=1}^\new{l} \|\mathcal{F}_p(\PINNpW,\new{t_i})\|^2 \, .
\end{equation}

\begin{figure}[h]
\begin{tikzpicture}[scale=1.0]
\node (A) at (0,-1) [circle,minimum size=1cm,fill=black!20,draw] {$t$};
\node (B) at (1.5,0) [circle,fill=black!20,minimum size=0.4cm,draw] {};
\node (C) at (1.5,-1) [circle,fill=black!20,minimum size=0.4cm,draw] {};
\node (D) at (1.5,-3) [circle,fill=black!20,minimum size=0.4cm,draw] {};
\node (I1) at (3,0) [circle,fill=black!20,minimum size=0.4cm,draw] {};
\node (I2) at (3,-1) [circle,fill=black!20,minimum size=0.4cm,draw] {};
\node (I3) at (3,-3) [circle,fill=black!20,minimum size=0.4cm,draw] {};
\node (E) at (4.5,0) [circle,fill=black!20,minimum size=0.4cm,draw] {};
\node (F) at (4.5,-1) [circle,fill=black!20,minimum size=0.4cm,draw] {};
\node (G) at (4.5,-3) [circle,fill=black!20,minimum size=0.4cm,draw] {};
\node (intermediate) at (3,1) {hidden layers};
\node (P) at (6,-1) [circle,minimum size=1cm,fill=black!20,draw] {$\!\!\mathcal{K}_p(t)\!\!$};
\node (R) at (9,0) [circle,minimum size=1cm,fill=black!20,draw] {$\!\scriptstyle\MSEU\!$};
\node (T) at (9,-2) [circle,minimum size=1cm,fill=black!20,draw] {$\!\scriptstyle\MSEF\!$};
\node (out) at (6,1) [align=center] {output\\[-0.25cm]layer};
\node (in) at (0,1) [align=center] {input\\[-0.25cm]layer};
\node (Vdots3) at (1.5,-2) [align=center] {$\vdots$};
\node (Vdots5) at (4.5,-2) [align=center] {$\vdots$};
\draw[->] (A) to (B);
\draw[->] (A) to (C);
\draw[->] (A) to (D);
\draw[->] (B) to (I1);
\draw[->] (B) to (I2);
\draw[->] (B) to (I3);
\draw[->] (C) to (I1);
\draw[->] (C) to (I2);
\draw[->] (C) to (I3);
\draw[->] (D) to (I1);
\draw[->] (D) to (I2);
\draw[->] (D) to (I3);
\draw[->] (I1) to (E);
\draw[->] (I1) to (F);
\draw[->] (I1) to (G);
\draw[->] (I2) to (E);
\draw[->] (I2) to (F);
\draw[->] (I2) to (G);
\draw[->] (I3) to (E);
\draw[->] (I3) to (F);
\draw[->] (I3) to (G);
\draw[->] (E) to (P);
\draw[->] (F) to (P);
\draw[->] (G) to (P);
\draw[->,dashed] (P) to node[above,sloped]{{\footnotesize training data}} (R);
\draw[->,dashed] (P) to node[below,sloped]{$\frac{d\mathcal{K}(t)}{dt}$} (T);
\end{tikzpicture}
\caption{\new{Network architecture of the proposed PINN. Each hidden layer consists of 30 neurons.}}
\label{fig:networkarch}
\end{figure}
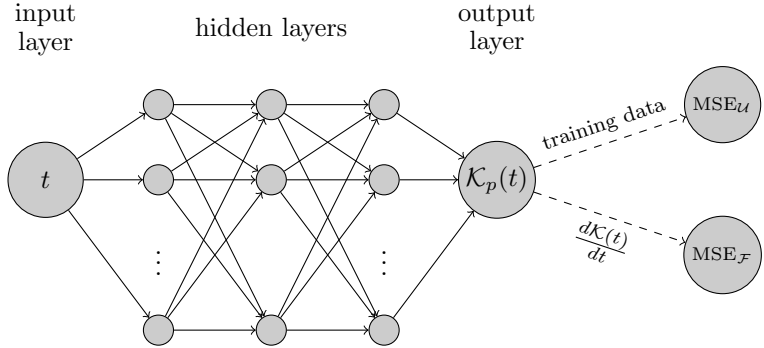

We introduce a hyperparameter and weighting factor $\alpha\in[0,1]$ weighting the data loss and residual loss in the loss function. 
We define the overall loss function as  
\begin{equation}\label{eq:alpha_opt}
    \mathcal{L}_\alpha = \mathcal{L}_\alpha(W)
    :=\alpha \, \MSEU + (1-\alpha) \, \MSEF
\end{equation}
and the minimization problem of the neural network as 
\begin{equation}\label{eq:loss_function_alpha}
    \underset{W}{\min} \bigl(\mathcal{L}_\alpha \bigr) \, .
\end{equation} 
\new{We want to note briefly that small values of the residual loss $\MSEF$ do not necessarily mean that the obtained solution is close to the exact solution of \eqref{eq:table1}. 
For this, one additionally needs the well-posedness of the problem \eqref{eq:table1} and the stability of the scheme used.}
% NETWORK IMPLEMENTATION / STRUCTURE  
For all implementations, the programming language \textit{Python} and the deep-learning framework of \textit{PyTorch} are used. 
\new{We used a standard network architecture with three fully connected hidden layers, each with 30 neurons, see Figure~\ref{fig:networkarch} for illustration. 
For each time point $t\in[0,T]$, the network returns values for all five compartment sizes in the vector $\mathcal{K}_p(t)\in\R^5$. 
The network architecture was inspired by Ben Moseley's harmonic oscillator PINN, see \cite{benmoseley}, and represents a reasonable compromise in terms of network complexity.}

\new{We used \new{hyperbolic tangent} activation functions and} the Adam optimizer for the training. 
We refer to \cite{ruder2016} for a survey of descent methods in machine learning.
\new{We also used a learning rate schedule that significantly reduces the learning rate after about $50\,\%$ of training iterations. 
More specifically, the learning rate $t(\kappa)$ in iteration $\kappa \in\{1,\dots,\kappa_{\text{max}}\}$ is given by 
\begin{equation}\label{eq:learningrate}
t(\kappa)= -( t_{\text{start}}-t_{\text{end}}) \,
\frac{\exp \bigl(\frac{\kappa - 0.5 \cdot \kappa_{\text{max}}}{0.08 \cdot \kappa_{\text{max}} } \bigr)}%
     {\exp \bigl(\frac{\kappa - 0.5 \cdot \kappa_{\text{max}}}{0.08 \cdot \kappa_{\text{max}} }\bigr) +1} +  t_{\text{start}},
\end{equation} 
where $t_{\text{start}}$ and $t_{\text{end}}$ denote the initial and final learning rates, respectively. 
In our experiments, we used $t_{\text{start}} = 0.003$ and $t_{\text{end}} = 0.00015$. 
Figure~\ref{fig:lrs} shows the learning rate schedule for $\kappa_{\text{max}} = 1000$ iterations.
\begin{figure}[htpb!] \centering
    \includegraphics[width=8cm]{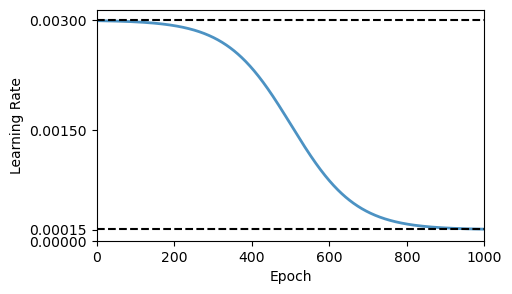}
    \caption{\new{Our employed learning rate schedule reducing the learning rate over 1000 epochs.}}   
\label{fig:lrs}
\end{figure}
}

% DATA %%%%%%%%%%%%%%%%%%%%
The data loss was evaluated with data sets obtained from the RKI \cite{rki1, rki2} and the German Vaccination Dashboard \cite{impfdashboard}. 
 These data were preprocessed \new{and rescaled to unit intervals before calculating the data loss $\MSEU$. 
Since this leads to a corresponding rescaling of the PINN predictions $\mathcal{K}_p$, the corresponding scaling factors were also included in the ODE system \eqref{eq:table1} when evaluating $\MSEF$.}

The data refers to the calendar weeks 10 in 2020 to 14 in 2022. Weekly case-hospitalization, 
case-fatality and vaccination rates were computed based on the given data sets. 
The RKI registers deceased individuals, in whom the SARS-CoV-2 pathogen was detected, as people who died from COVID-19.

\new{The weighted loss function $\mathcal{L}_\alpha=\alpha \MSEU + (1-\alpha) \MSEF$ consists of the data loss and the residual loss term. 
As stated in \cite{karniadakis1}, training using the data loss (i.e., measurements, physics-uninformed) is considered supervised learning, while training with the residual loss using the governing differential equation (physics-informed) is considered unsupervised learning.}

%%%%%%%%%%%%%%%%%%%%%%%%%%%%%%%%%%%%%%%%%%
\subsection{Conflicting Training Goals: Pareto Front and Trade-Off Analysis}\label{subsec:paretofront}
In this paper, we take a biobjective perspective on the optimization problem described in \eqref{eq:alpha_opt}. 
This is the basis for a thorough trade-off analysis regarding the two loss terms $\MSEU$ and $\MSEF$. 
Rather than considering a weighted sum of these two training goals with a fixed weighting parameter \new{$\alpha\in[0,1]$}, we consider both optimization goals independently and comprise them in a vector-valued objective function $\mathcal{L}$ that maps every feasible solution vector $(W,p)$ to a two-dimensional outcome vector $y=\mathcal{L}(W,p)\in\R^2$:
\begin{equation}\label{eq:optproblem}
    \underset{W}{\min} \, \mathcal{L}_{\alpha}(W)
    =\underset{W}{\min} \, \bigl( \MSEU(W),\MSEF(W)\bigr) .
\end{equation}
As before, $W$ denotes the neural network weights. 
To analyze the biobjective optimization problem~\eqref{eq:optproblem}, we first review some basic concepts from the field of multiobjective optimization. For a more detailed introduction into this field see, e.g., \cite{ehrgott05multicriteria}.

Towards this end, we denote by $Y$ the \textit{outcome set of problem \eqref{eq:optproblem}} that includes %denotes 
all possible outcome vectors $y=\mathcal{L}_{\alpha}(W)\in\R^2$. 
A solution $\hat{W}$ \new{(i.e., a set of NN-weights $\hat{W}$ resulting from the training)} is called \textit{efficient} or \textit{Pareto optimal} if there exists no other solution $W$ \new{(i.e., no other set of NN-weights $W$)} such that $\MSEU(W)\le \MSEU(\hat{W})$ and $\MSEF(W)\le \MSEF(\hat{W})$, where at least on of these two inequalities is strict. 
If $\hat{W}$ is Pareto optimal, then the corresponding outcome vector $\new{y=}(\MSEU(\hat{W}),\MSEF(\hat{W}))$ is called \textit{nondominated point}. 
Hence, Pareto optimal solutions are those solutions \new{(i.e., NN weights)} that can not be improved in one loss function without deterioration in the other loss function. %w.r.t.\ both loss functions simultaneously.
The set of all Pareto optimal solutions (nondominated points, respectively) is denoted by $X_E$ ($Y_N$, respectively). 
Note that $Y_N$ is also often referred to as \textit{Pareto front}. \new{The ultimate goal is the efficient, i.e., fast approximation of so-called \textit{knee solutions} that provide near-optimal values for both data loss \textit{and} residual loss.
This is realized by a dichotomic search strategy specifically tailored to NN training.}

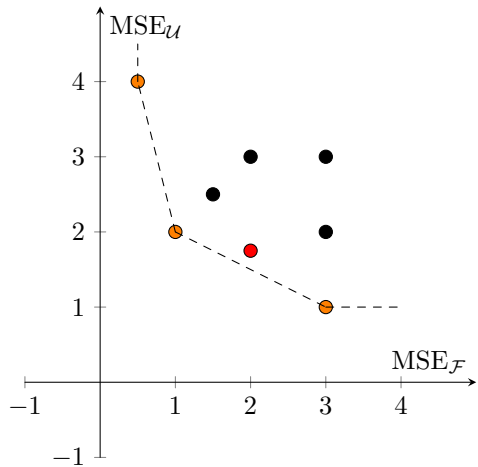
\begin{figure}[!htpb]
\centering
\begin{tikzpicture}[line cap=round,line join=round,>=triangle 45,x=1cm,y=1cm]
\begin{axis}[
x=1cm,y=1cm,
axis lines=middle,
xlabel = $\MSEF$,
ylabel = $\MSEU$,
xmin=-1,
xmax=5,
ymin=-1,
ymax=5,
xtick={-1,0,1,...,4},
ytick={-1,0,1,...,4},]
\clip(0,0) rectangle (5,5);
\begin{scriptsize}
\draw [fill=orange] (0.5,4) circle (2.5pt);
\draw [fill=orange] (1,2) circle (2.5pt);
\draw [fill=red] (2,1.75) circle (2.5pt);
\draw [fill=orange] (3,1) circle (2.5pt);
\draw [fill=black] (1.5,2.5) circle (2.5pt);
\draw [fill=black] (2,3) circle (2.5pt);
\draw [fill=black] (3,2) circle (2.5pt);
\draw [fill=black] (3,3) circle (2.5pt);
\draw[dashed] (1,2) to (3,1);
\draw[dashed] (1,2) to (0.5,4);
\draw[dashed] (0.5,4) to (0.5,4.5);
\draw[dashed] (3,1) to (4,1);
\end{scriptsize}
\end{axis}
\end{tikzpicture}
\caption[Nondominated Points]
{\new{Schematic illustration of a potential set of outcome vectors $(\MSEF,\MSEU)$ obtained from NN training,} Points belonging to $Y_{\new{s}N}$ are depicted in orange, \new{the unsupported nondominated point is shown in red. The dashed line outlines the convex hull of the set of outcome vectors $Y+\R^2_{\geq}$.}} 
\label{fig:nondompoints}
\end{figure}

\new{An approximation of the Pareto front yields an approximation of the knee solution. It can be found by} 
solving a series of parametric single-objective subproblems, so-called \textit{scalarizations} (see again, e.g., \cite{ehrgott05multicriteria}). 
To keep these subproblems simple, we use a \textit{weighted sum} approach that leads to the single-objective optimization problem~\eqref{eq:loss_function_alpha} with the objective function \eqref{eq:alpha_opt}, i.e.,  
\begin{equation}\tag{\ref{eq:loss_function_alpha}}\label{eq:weightedsum}
      \underset{W}{\min} \bigl(\mathcal{L}_\alpha(W) \bigr) = \underset{W}{\min} \bigl\{ \alpha \, \MSEU(W) + (1-\alpha) \, \MSEF(W) \bigr\}
\end{equation} 
It is a well-known fact \cite{ehrgott05multicriteria} that for all weighting parameters $\alpha\in (0,1)$, an optimal solution of \eqref{eq:weightedsum} is always part of $Y_N$. 
However, the converse statement % that all solutions of $Y_N$ can be found by solving \eqref{eq:weightedsum} with varying $\alpha$ 
is only true under convexity assumptions. 
Indeed, the complete set $Y_N$ can be generated by varying the weighting parameter $\alpha\in(0,1)$ whenever the set $Y$ is $\R^2_\ge$-convex, a property that can generally not be guaranteed in neural network training.
We recall that Pareto optimal solutions  $\hat{W}\in X_E$ are called \textit{supported} if there is some $\alpha\in(0,1)$ such that $\hat{W} \in X_E$ is an optimal solution of \eqref{eq:weightedsum}.  
The sets of all supported efficient solutions and supported nondominated points are denoted $X_{sE}$ and $Y_{sN}$, respectively. 
Note that all supported nondominated points are located on the boundary of the convex hull of $Y$ (see, e.g., \cite{przybylski2019simple}).
Figure~\ref{fig:nondompoints} shows an example of a set of \new{supported} nondominated points $Y_{\new{s}N}$ within a set of outcome vectors $Y$ in $\R^2$ \new{as well as an unsupported nondominated point}. 
\new{The choice of the weighting parameter $\alpha$ in \eqref{eq:weightedsum} is both a crucial and highly challenging task. 
While a too large value of $\alpha$ overemphasizes data loss and therefore often leads to overfitting, a too small value of $\alpha$ may give too much priority to a possibly only approximate physical model.} 
When trying to approximate the Pareto front by scanning over potential weights $\alpha\in(0,1)$, even for biobjective and convex problems predefined weighting parameters may lead to very un-evenly distributed points on the Pareto front, see, e.g., \cite{das1997}. 
Moreover, this approach does not scale well to higher-dimensional problems that include more than two training objectives. 
\new{Indeed, when searching through all reasonable weights, the required number of weighting vectors $\alpha$ (and hence the individual NN training with respect to a weighted sum objective \eqref{eq:weightedsum}) grows exponentially with the number of objective functions.}

In the next section %\ref{subsec:dichotomic} 
we present an \new{efficient} approach that supports an adaptive selection of weighting parameters \new{to identify knee solutions for the case of two optimization goals (the data loss and the residual loss in our case). 
We emphasize that our approach can be generalized to more than two objective functions (e.g.\ for the PINN solution of PDEs) and thus has the potential to significantly reduce the number of individual NN trainings required to identify near-ideal NN weights.}

%%%%%%%%%%%%%%%%%%%%%%%%%%%%%%%%%%%%%%%%%
\subsection{Dichotomic Search for Adaptive Pareto Front Approximations}\label{subsec:dichotomic}
\new{Near-ideal NN-solutions can be identified by adapting the} % achieved 
scalarization-based dichotomic search algorithm described in \cite{przybylski2019simple}. \new{The idea is to compute adaptively weighting parameters $\alpha$ that refine the current approximation of the Pareto front in the most promising regions.} 
This approach aims at an automatic adaptation to the curvature and scaling of the problem in order to quickly find a diverse set of solutions, \new{and to quickly identify near-ideal knee solutions.} 
Moreover, dichotomic search can be easily integrated in an interactive procedure that allows to zoom in into specific parts of the Pareto front that are most interesting to the decision maker, see, e.g., \cite{klam:inte:2008}.

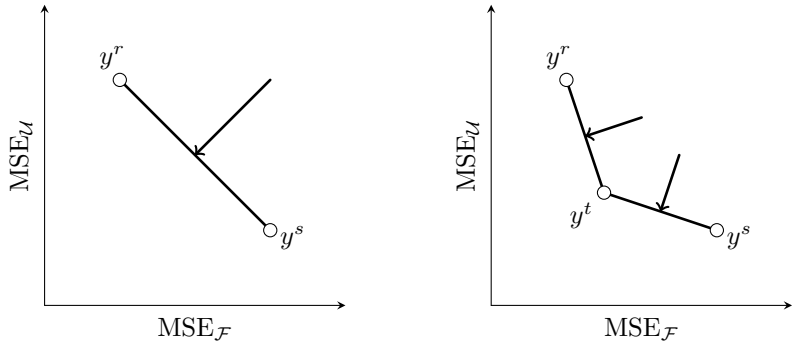
\begin{figure}[!htbp]
\centering
\begin{minipage}[t]{0.48\linewidth}
\centering
\begin{tikzpicture}[line cap=round,line join=round,>=triangle 45,x=1cm,y=1cm]
\begin{axis}[
x=1cm,y=1cm,
axis lines=middle,
%ymajorgrids=true,
%xmajorgrids=true,
x label style={at={(axis description cs:0.5,-0.01)},anchor=north},
y label style={at={(axis description cs:-0.01,0.5)},rotate=90,anchor=south},
xlabel = $\MSEF$,
ylabel = $\MSEU$,
xmin=0,
xmax=4,
ymin=0,
ymax=4,
xtick={0},
ytick={0},]
\clip(0,0) rectangle (4,4);
\draw [name path = A, line width=1pt] (1,3)-- (3,1);
\draw [name path = B, line width=1pt, -to] (3,3)-- (2,2);
\node[] at (0.9,3.3) {$y^r$};
\node[] at (3.3,0.9) {$y^s$};
\begin{scriptsize}
\draw [fill=white] (1,3) circle (2.5pt);
\draw [fill=white] (3,1) circle (2.5pt);
\end{scriptsize}
\end{axis}
\end{tikzpicture}
\end{minipage}
\begin{minipage}[t]{0.48\linewidth}
\centering
\begin{tikzpicture}[line cap=round,line join=round,>=triangle 45,x=1cm,y=1cm]
\begin{axis}[
x=1cm,y=1cm,
axis lines=middle,
%ymajorgrids=true,
%xmajorgrids=true,
x label style={at={(axis description cs:0.5,-0.01)},anchor=north},
y label style={at={(axis description cs:-0.01,0.5)},rotate=90,anchor=south},
xlabel = $\MSEF$,
ylabel = $\MSEU$,
xmin=0,
xmax=4,
ymin=0,
ymax=4,
xtick={0},
ytick={0},]
\clip(0,0) rectangle (4,4);
\draw [name path = A, line width=1pt] (1,3)-- (1.5,1.5);
\draw [name path = B, line width=1pt, -to] (2,2.5)-- (1.25,2.25);
\draw [name path = A, line width=1pt] (1.5,1.5)-- (3,1);
\draw [name path = B, line width=1pt, -to] (2.5,2)-- (2.25,1.25);
\node[] at (0.9,3.3) {$y^r$};
\node[] at (3.3,0.9) {$y^s$};
\node[] at (1.2,1.2) {$y^t$};
\begin{scriptsize}
\draw [fill=white] (1,3) circle (2.5pt);
\draw [fill=white] (3,1) circle (2.5pt);
\draw [fill=white] (1.5,1.5) circle (2.5pt);
\end{scriptsize}
\end{axis}
\end{tikzpicture}
\end{minipage}
\caption{%Mode of action of the dichotomic search. 
On the left: The connecting line segment between $y^r$ and $y^s$ is perpendicular to the vector $(\alpha,1-\alpha)^\top$ that is here multiplied with a negative scalar to show the minimization direction.
On the right: 
Since $(\alpha, 1-\alpha)^\top y^t < (\alpha, 1-\alpha)^\top y^r$, the new point $y^t$ is nondominated and two new subproblems are generated, see \cite[p. 7]{przybylski2019simple} (own illustration).
\label{fig:dichotomic}}
\end{figure}

The following ideas can be found in \cite{przybylski2019simple}. 
In the biobjective case, the dichotomic search comes down to solving a sequence of weighted sum scalarizations \eqref{eq:weightedsum} with $\alpha\in(0,1)$ and makes use of the fact that in the two-dimensional case, for two nondominated points $y^r$ and $y^s$ it holds that  $y_1^r < y_1^s$ implies $y_2^r > y_2^s$. 
A weighted sum scalarization \eqref{eq:weightedsum} with 
$\alpha = (y_2^r-y_2^s) / c > 0$ and $1-\alpha = (y_1^s-y_1^r) / c > 0$, where $c = y^r_2-y^s_2+y^s_1-y^r_1$, is then solved to find new supported points  between $y^r$ and $y^s$. 
The weighting parameter $\alpha$ hence defines a normal vector to the line segment connecting $y^r$ and $y^s$ since
\begin{align*}
    \bigl(y^r-y^s\bigr)^\top \begin{pmatrix}
    \alpha\\ 1-\alpha
    \end{pmatrix} &= 
    \begin{pmatrix}
    y_1^r - y_1^s, & y_2^r - y_2^s
    \end{pmatrix}
    \begin{pmatrix}
    \alpha\\ 1-\alpha
    \end{pmatrix}\\
    &=     
    c\,\cdot\begin{pmatrix}
    -(1-\alpha), & \alpha
    \end{pmatrix}
    \begin{pmatrix}
    \alpha\\ 1-\alpha
    \end{pmatrix} = 0.
\end{align*}
This is illustrated in the left of Figure~\ref{fig:dichotomic}. %with $\alpha$ being multiplied by a negative scalar to emphasize the minimization sense.
Solving the weighted sum problem with the new weighting parameter $\alpha$ leads to a nondominated point $y^t$ (if the problem is solved to global optimality) for which two cases can occur:
\begin{enumerate}
    \item If $(\alpha,1-\alpha)^\top y^t < (\alpha,1-\alpha)^\top y^r$, then $y^t$ is a new supported nondominated point. 
    Two new subproblems are generated, one of which is defined by $y^r$ and $y^t$ while the other one is defined by $y^t$ and $y^s$. 
    This case is illustrated in the right of Figure~\ref{fig:dichotomic}.
    \item If $(\alpha,1-\alpha)^\top y^t = (\alpha,1-\alpha)^\top y^r=(\alpha,1-\alpha)^\top y^s$, then $y^t$ lies on the line segment connecting $y^r$ and $y^s$ and the search can stop in this interval.
\end{enumerate}
The dichotomic search progresses in \textit{levels}, where level~1 contains the outcome vectors of the weighted sum scalarization with the two initial weights $\alpha_1$ and $\alpha_2$, $\alpha_1<\alpha_2$, and one dichotomy step (see the left part of Figure~\ref{fig:dichotomic} for an illustration of the associated weighting parameter). In level~2, weighted sum scalarizations are solved for all weights defined by the line segments comprising the convex hull of the current approximation of the Pareto front (see the right part of Figure~\ref{fig:dichotomic} for an illustration).
The search is repeated until a predefined number of levels has been evaluated, or until no new subproblems have been generated.

Algorithm~\ref{alg:dichotomic} summarizes the implementation of the \new{bisection enhanced} dichotomic search (\new{BEDS}) that \new{enhances the dichotomic search scheme by an occasional bisection step}.
It is based on \cite{reiners} 
\new{and considers
% takes into account 
the fact that NN training with a certain weight parameter $\alpha$ may yield a local minimum, i.e., the training 
%(in our case the Adam optimizer) 
(here the Adam optimizer) 
may terminate in a local minimum and not in a global minimum as assumed in the general dichotomic search procedure.} 

\begin{algorithm}[h!tb]
 \KwData{Training data, hyperparameter settings for \new{Adam} solver, depth of the search (levels), initial weighting parameters $\alpha_1,\alpha_2\in(0,1)$, $\alpha_1<\alpha_2$, to approximate extremal solutions focusing on $\MSEU$ and $\MSEF$, respectively (see \eqref{eq:alpha_opt})}
 \KwResult{Approximation of the Pareto front and corresponding PINN parameters} %$w^*$ approximating a Pareto knee}
 $\Lambda \leftarrow \{ \alpha_1,\alpha_2 \}$\;
 $\mathtt{cand} \leftarrow \emptyset$\;
 \For{$l=1,\ldots,\mathtt{levels}$}{
    \For{$\alpha \in \Lambda$}{
      train with weighted sum objective $\mathcal{L}_{\alpha}$\; 
      add objective vector \new{$(\MSEU,\MSEF)$} to $\mathtt{cand}$\;
      }
    delete all dominated points in $\mathtt{cand}$\;
    sort $\mathtt{cand}$ by second objective function (in increasing order)\;
    \If{$l<\mathtt{levels}$}{
      \For{$i\in\{2,\ldots,\vert \mathtt{cand}\vert\}$}{
        $\text{diff} \leftarrow \mathtt{cand}(i) - \mathtt{cand}(i-1)$\;
%         $\alpha_{\text{new}} \leftarrow \frac{-\text{diff}_1}{\text{diff}_2 -\text{diff}_1}$ and 
%         add $\alpha_{\text{new}}$ to $\mathtt{list}$\;
        \(\alpha_{\text{new}} \leftarrow \Bigl\{ \frac{\text{-diff}_1}{\text{diff}_2 -\text{diff}_1} \Bigr\}\)\;
        \If(\tcp*[f]{failed weight setting}){\new{$(0.999-\alpha_{\text{new}}) < 0.001$ \upshape{\textbf{or}} $\alpha_{\text{new}} < 0.8$}}{
            \new{$\alpha_{\text{new}} \leftarrow(\text{succ}(\alpha) + \text{pred}(\alpha))/2$};\\  \tcp*[f]{replace unsucc.\ $\alpha$ by bisecting parent interval}\;
        }
        \(\Lambda \leftarrow \Lambda \cup \alpha_{\text{new}}\)\;
      }
      resort \(\Lambda\) increasingly
    }
 }
 Return nondominated points from $\mathtt{cand}$ to illustrate trade-offs\;
 \caption{\new{Bisection Enhanced} Dichotomic Search \new{(BEDS)} \label{alg:dichotomic}}
\end{algorithm}

\new{To overcome the numerical difficulties arising from this fact, the dichotomic scheme is enhanced by bisection steps that generate new - and promising - weighting parameters $\alpha$ when needed.
The bisection step occurs whenever the newly found weight falls outside the original search interval $[\alpha_1, \alpha_2]$. 
This can happen if the calculated slope of the dichotomic search becomes too small, leading to numerical problems.
Algorithm~\ref{alg:dichotomic} performs several full training runs, resetting the weights of the network after line 6.
The variable $\mathtt{cand}$ is initialized with an empty set and stores all objective vectors $y=(\MSEU,\MSEF)$ generated during training for different weighting parameters $\alpha$ of the loss function $\mathcal{L}_{\alpha}$.}

%%%%%%%%%%%%%%%%%%%%%%%%%%%%%%%%%%%%%%%%%%%%%%%%%%%%%%%%%%%%%%%%%%% 
\section{Results}\label{sec:results}
This section is subdivided into three parts. 
In Section~\ref{subsec:valid_short}, our PINN is validated.
\new{We first use the data generated during the delta variant ($85^{th}$ to $100^{th}$ considered week or week 41/2021 to week 4/2022) and predict the first omicron wave ($100^{th}$ to $104^{th}$ considered week or week 4/2022 to week 8/2022).
We use the NSFD scheme to determine fitting parameter values for $\beta$ and $\kappa$.}

\new{In Section~\ref{subsec:dichotomicresults}, we present the results of dichotomic search to investigate the impact of the weighting parameter $\alpha$ on the training objective \eqref{eq:alpha_opt}. 
We discuss the ability of dichotomic search to approximate a Pareto front, which can help in finding reasonable trade-off solutions.}

\new{Finally, in Section~\ref{subsec:valid_long}, we present a result using a long training dataset, starting from the beginning of the pandemic in January 2020. 
The differences between the results obtained with the two training datasets are discussed. 
Thus, we distinguish between training datasets that cover the time since the outbreak of the pandemic in Germany, 
which we refer to as \textit{long-term training data}, and training datasets that contain only data from a particular wave or even a peak reached during the pandemic, which we refer to as \textit{short-term training data}.}

%

%%%%%%%%%%%%%%%%%%%%%%%%%%%%%%%%%%%%%%%%%%%%%%
\subsection{Validation of Scenarios Generated with Short-Term Training Data}\label{subsec:valid_short}

\new{In this part, we use short-term training data and use it to predict the infected compartment $I$ for the following weeks. 
The reason we use short-term data is that our physical model is built to describe the behavior of exactly one wave of infection with one maximum. 
Since there were many waves of different virus variants described by different infection parameters in the COVID-19 pandemic, using only one wave for training to predict the next wave is a more realistic approach to incorporate our model. 
Nevertheless, we will present results with a longer training period of training data for comparison in Section~\ref{subsec:valid_long}. 
It is important to note that data covering the increase in infection rates due to omicron spread must be included in the training data to predict the further increase. 
The PINN adjusts its trained parameters to the underlying data set.}

\begin{figure}[htpb!] \centering 
     \includegraphics[width=1.0\textwidth]{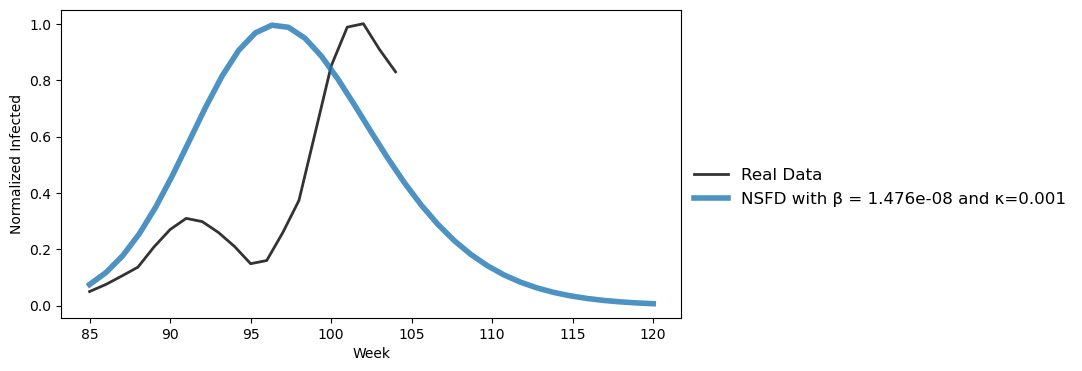}  
     \caption{\new{Normalized infection numbers from the $85^{th}$ considered week to the $104^{th}$ considered week and the normalized infected part of the NSFD scheme.}}
\label{fig:nsfd_determination}
\end{figure}

\new{We first use the NSFD scheme to determine appropriate parameter values for the infection rates $\beta$ and $\kappa$. Figure~\ref{fig:nsfd_determination} shows (normalized) infected data in the time frame we use to train and validate our PINN, along with the predictions we obtained using the NSFD scheme (see Eq.~\eqref{eq:table2}) with the parameters in the right column of Table~\ref{table:parameters}, i.e., $\beta = 0.00000001476$ and $\kappa = 0.001$.}

\new{We see that this parameter choice reproduces well both the initial slope of the delta wave and the maximum of the omicron wave. 
The NSFD scheme with a constant transmission risk obviously predicts one maximum in the course of one wave. 
We note that the inclusion of a time-varying transmission rate in the NSFD approach would facilitate the prediction of multiple peaks within a wave, as shown in \cite{treibert22}. 
For simplicity, we have not included a time-varying rate in the PINN or NSFD scheme used in this work; this is left for future research.}

\begin{figure*}[htpb!]
\subfigure[\new{Perfect fit of training data, weak prediction.}\label{fig:validnewdata1_1}]
{\includegraphics[width=1.0\textwidth,clip]{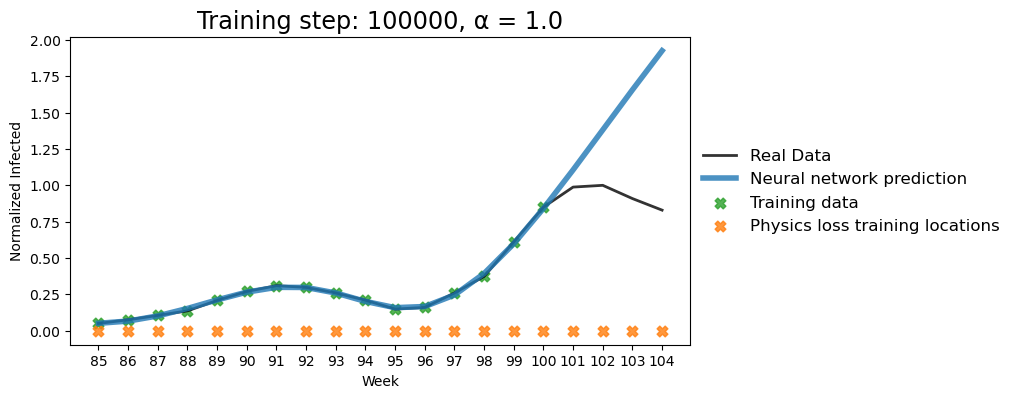}}
\\
\subfigure[\new{Good fit of training data, good prediction.} \label{fig:validnewdata1_2}]
{\includegraphics[width=1.0\textwidth,clip]{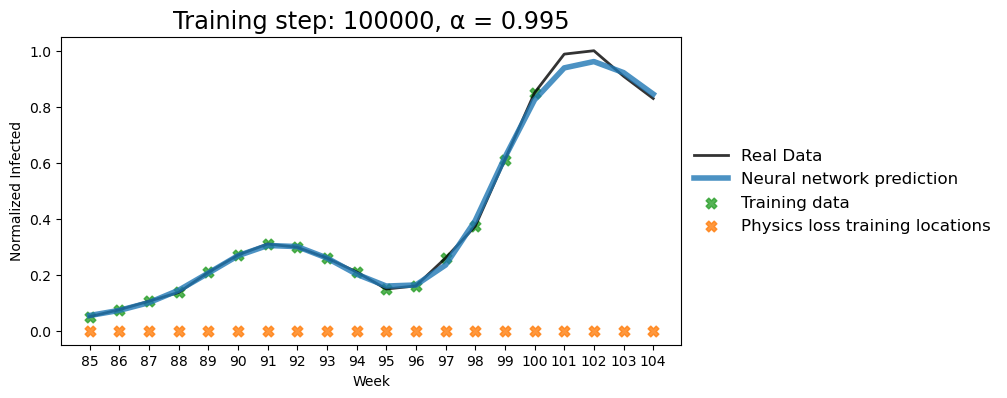}}
\\
\subfigure[\new{The prediction gets worse if $\MSEF$ is weighted too heavily.} \label{fig:validnewdata1_3}]
{\includegraphics[width=1.0\textwidth,clip]{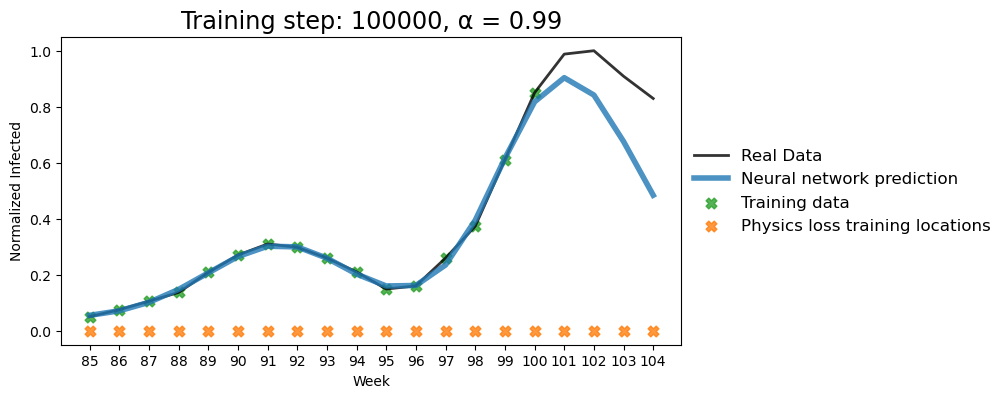}}
\\
\caption{\new{Infection numbers obtained from the reported data (black) or the training of the PINN with the loss terms $\MSEU$ and $\MSEF$ (blue) from the 85$^{th}$ considered week to the 104$^{th}$ week for three different assignments of the weighting parameter $\alpha$. 
The training data set covers the calendar weeks 41 in 2021 to 4 in 2022 ($85^{th}$ to $100^{th}$ considered week).}}
\label{fig:validnewdata1}
\end{figure*}

\new{Figure~\ref{fig:validnewdata1} shows the results of a short-term prediction in which the PINN was trained on data ranging from the $85^{th}$ week under consideration to the $100^{th}$ week under consideration and used to predict the following four weeks (weeks $101-104$).
We used three different values for the weighting parameter $\alpha \in \{1.0,0.995,0.99\}$, see Eq.~\eqref{eq:alpha_opt}. 
Each training run included $100000$ iterations.}

\new{We note that the weighting parameter $\alpha$ plays an important role in the quality of the prediction. 
In Figure~\ref{fig:validnewdata1_1}, $\alpha = 1$ was chosen, so the loss only considered $\MSEU$. 
In this case, PINN succeeds in fitting the training data almost perfectly, but it fails to predict the decline in infection numbers in the following weeks because it has no information about the physical properties of the system, leading to overfitting. 
Figure~\ref{fig:validnewdata1_2} shows that using $\alpha = 0.995$ leads to a slightly less perfect fit to the training data, but a good prediction of the infection numbers in the following four weeks.
Finally, Figure~\ref{fig:validnewdata1_3} shows that overweighting $\MSEF$ again leads to suboptimal predictions, which can be explained by the fact that the PINN has learned the physical parameters of the delta wave, which are different from those of the omicron wave. 
It can therefore be observed that the data loss must be weighted much higher than the residual loss to obtain a good prediction in applications where the physical model only partially explains the true dynamics}. 

\new{However, complete neglect of residual loss still does not lead to the best performance of our PINN. 
Although no mathematical model can optimally describe an infectious disease because not every epidemiological detail relevant to transmission is known, we can use residual loss to incorporate systematic knowledge about the spread and transmission dynamics of the disease into our neural network. 
Since $\alpha=1$ does not yield the smallest errors in our validation runs, the inclusion of the residual error is still justified and reasonable. 
In general, predictions of future pandemic dynamics due to unknown mutational variants are subject to many uncertainties, changing intervention measures or compliance of the population and new vaccination strategies.} 

\new{Table~\ref{tab:errorvalues2} shows the mean squared error between the infected compartment of our PINN prediction $\mathcal{K}_p^3(t_i)$ and the reported infection data $\hat{\mathcal{K}}_p^3(t_i)$ in the weeks $85$ to $104$ considered (i.e., for the entire time frame for which data were available, not just the time frame used for training), where the mean squared error is calculated as follows}
\begin{equation}\label{eq:valerror}
  \new{\text{MSE}_{\text{val}} := \frac{1}{l}\sum_{i=1}^l \|\mathcal{K}_p^3(t_i)-\hat{\mathcal{K}}_p^3(t_i)\|^2\,.}
\end{equation}

\begin{table}
\begin{tabular}{l|lllllll}
\textbf{Value of} $\alpha$ & 1.0    & 0.999  & 0.998  & 0.997  & 0.996  & 0.995  & 0.994  \\
$\text{MSE}_{\text{val}}$      & 0.0961 & 0.0553 & 0.0505 & 0.0031 & 0.0012 & 0.0003 & 0.0024 \\ \hline
\textbf{Value of} $\alpha$ & 0.993  & 0.992  & 0.991  & 0.99   &   0.95   &   0.9    &   0.8     \\
$\text{MSE}_{\text{val}}$       & 0.0016 & 0.0097 & 0.0109 & 0.0104 & 0.0772 & 0.0853 & 0.1030      
\end{tabular}
\caption{\new{Mean squared errors between the reported infection data and the PINN results using the $85^{th}$ to $104^{th}$ considered week, depending on the weighting parameter $\alpha$. 
This can be considered a validation because it includes the error for the prediction for weeks 100 to 104 in addition to the data used for training (weeks 85 to 100).}}\label{tab:errorvalues2}
\end{table}
To create Table~\ref{tab:errorvalues2}, we manually modified the weighting parameter $\alpha\in(\new{0.8},1)$. It is remarkable that the smallest error is achieved with \new{$\alpha=0.995$} (among the considered parameter values.)
With \new{$\alpha=0.996$}, the error becomes \new{four} times as large and with \new{$\alpha=0.994$} it becomes \new{eight} times as large. 
Note, however, that the individual training runs do not necessarily terminate with a (globally) optimal solution; 
so the reported error values can only approximate the best possible error for the respective choices of weighting parameters.
Nevertheless, we can observe clear reductions in the performance of the network when setting $\alpha=0.9$ \new{or $\alpha=1.0$}. 
For instance, we obtain an approximate \new{eight-fold increase of $\text{MSE}_{\text{val}}$} if the weighting parameter $\alpha$ is decreased from \new{$\alpha=0.99$ to $\alpha=0.9$}. 

The above discussion shows that the choice of the weighting parameter $\alpha$ plays an important role in the prediction quality. 
Note that the prediction quality here is measured by comparisons with the real data. 
It is therefore not surprising that the prediction quality is higher when the data loss $\MSEU$ is highly weighted in the weighted sum training objective \eqref{eq:alpha_opt}. 
However, the results also show that the residual loss $\MSEF$ should not be ignored. 
This motivates a more detailed analysis of the trade-off between data loss and residual loss using dichotomic search in the following section.

%%%%%%%%%%%%%%%%%%%%%%%%%%%%%%%%%%%%%%%%%%%%%%
\subsection{Results of the Dichotomic Search}\label{subsec:dichotomicresults}
\begin{figure}[htbp!]
\subfigure[\new{Level 1: Lexicographic optimization and one weighted sum (three training runs)}\label{fig:knee_level1}]
{\includegraphics[width=0.48\textwidth,trim=10 0 37 39]{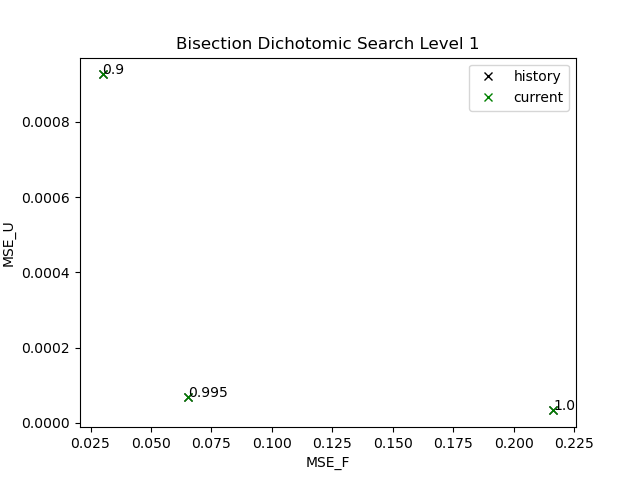}}
\hfill
\subfigure[\new{Level 2: Five training runs, bisection used}\label{fig:knee_level2}]
{\includegraphics[width=0.48\textwidth,trim=10 0 37 39]{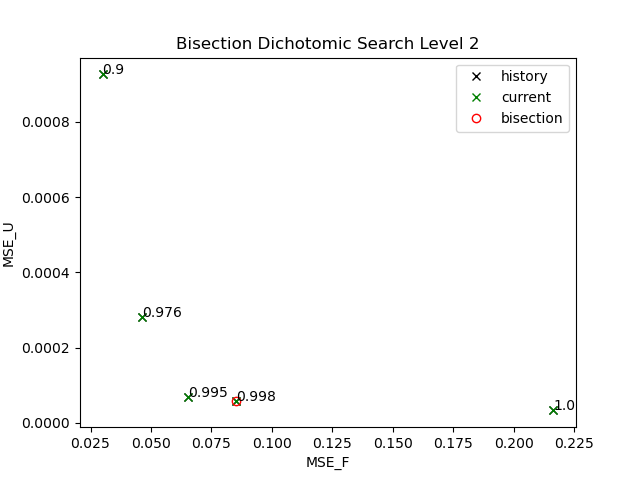}}
\vspace{0.5cm}
\\
\subfigure[\new{Level 3: Nine training runs} \label{fig:knee_level3}]
{\includegraphics[width=0.48\textwidth,trim=6 0 37 39]{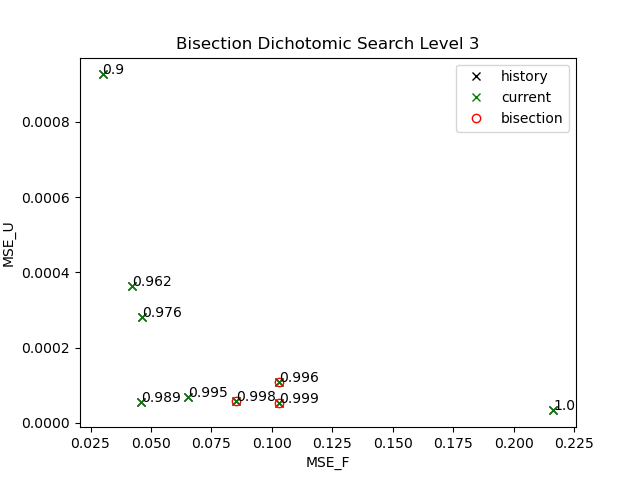}}
\hfill
\subfigure[\new{Level 4: Final Level with $13$ training runs} \label{fig:knee_level4}]
{ \includegraphics[width=0.48\textwidth,trim=6 0 37 39]{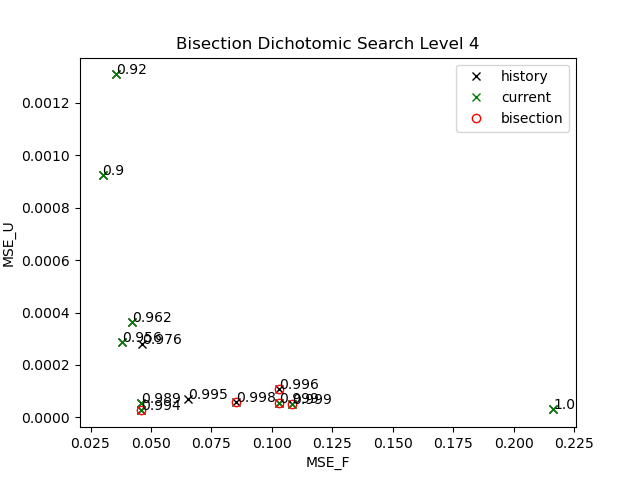}}
\caption{\new{Pareto front approximation using dichotomic search with a starting learning rate of 0.003 and learning rate schedule with starting weights $\alpha_1 = 0.9$ and $\alpha_2 = 0.999$. 
Favorable trade-offs %and knee solutions 
can be identified starting from Level~3.}\label{fig:knee}}
\end{figure}

In this section, we present the results of dichotomic search using Algorithm~\ref{alg:dichotomic}. 
Based on the discussion in the previous sections, we focus on short-term training data from calendar week \new{41} in 2021 to \new{4} in 2022 (\new{$85^{th}$ to $100^{th}$} considered week). 
See Section~\ref{subsec:valid_short} and Figure~\ref{fig:validnewdata1} for comparison. 
Four levels of Pareto front approximations (based on three, five, nine and 13 training runs, respectively) are shown in 
in Figure~\ref{fig:knee}. 
Each training run was performed  with \new{100000} epochs and the Adam optimizer with \new{a learning rate schedule as defined in equation~\eqref{eq:learningrate} with $t_{\text{start}} = 0.003$ and $t_{\text{end}} = 0.00015$.}
The search was initiated with \new{$\alpha_1 = 0.9$ and $\alpha_2 = 1.0$.}
Other search windows can be used depending on preferences or if additional information is available.
In addition, the search window can be used to zoom into a specific part of the Pareto front. 

The results confirm that a pronounced Pareto front with a diverse set of outcome vectors \new{and a clear knee solution} was approximated after only a few training runs. 
Indeed, level~2 based on five training runs (see Figure~\ref{fig:knee_level2}) already provides a rough approximation of the Pareto front. 
At level 3 (Figure~\ref{fig:knee_level3}), after nine training runs, we already have a very good approximation of the Pareto front.
The final level 4 is based on \new{13} training runs (Figure~\ref{fig:knee_level4}). 
On a machine with an AMD Ryzen 7 3700X 8-Core Processor and NVIDIA GeForce RTX 3060 graphic card, 
each training run took about 3 minutes, so the final level needed a computation time of about \new{39} minutes. 
The computation time of each training run is of course dependent on the specific problem and network, 
however, the goal here was to show that one can gain insight into the complete Pareto front of the problem with rather few training runs.

This shows that the dichotomic search scheme was successful in obtaining a diverse set of solutions with only very few training runs necessary, allowing the decision maker insight into the effect of changing the weighting parameter $\alpha$ that reflects the importance of each loss function.

\new{The results also show that it is possible to significantly improve the loss with respect to $\MSEF$ without losing much in the $\MSEU$ part of the loss function. 
While with $\alpha=1$ we obtain $\MSEF=0.2176$ and $\MSEU=7.495 \cdot 10^{-5}$, $\alpha=0.994$ leads to $\MSEF=0.0471$ and $\MSEU=7.599 \cdot 10^{-5}$.
We can therefore use the BEDS algorithm to quickly and automatically identify favorable trade-offs in the parameter weighting of PINN loss functions.}

Overall, it can be seen that the variation of $\alpha$ has a large impact on the relative importance of data loss and residual loss.

%%%%%%%%%%%%%%%%%%%%%%%%%%%%%%%%%%%%%%
\subsection{Validation of Scenarios Generated with Long-Term Training Data}\label{subsec:valid_long}
\begin{figure}[htpb!] \centering 
     \includegraphics[width=1.0\textwidth]{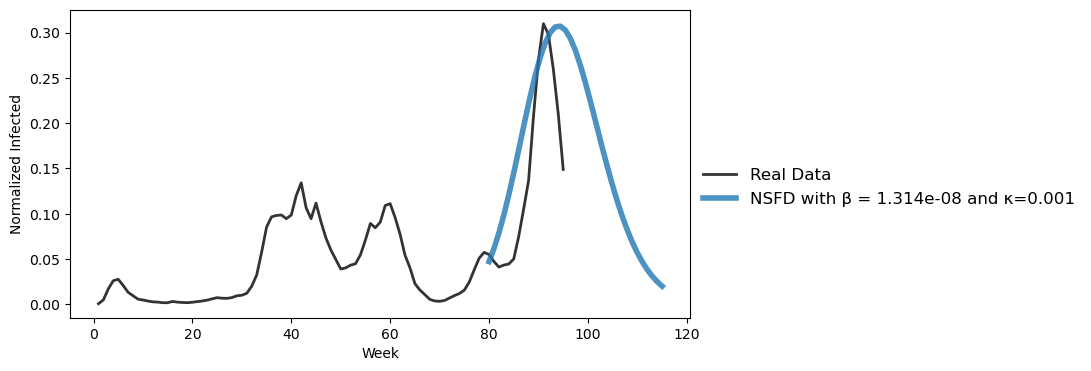}  
     \caption{\new{Normalized infection numbers from the $1^{st}$ considered week to the $95^{th}$ considered week and the normalized infected part of the NSFD scheme.}}
\label{fig:nsfd_determination_2}
\end{figure}

\new{Finally, we want to use our results to make a long-term prediction using most of our available data. 
We use the $1^{st}$ to $89^{th}$ week under consideration (week 10/2020 to week 45/2021) as our training data and let the PINN predict the following delta wave ($89^{th}$ to $95^{th}$ week under consideration or week 45/2021 to week 51/2021). 
First, we reapply the NSFD scheme to reevaluate the parameter values for $\beta$ and $\kappa$. Figure~\ref{fig:nsfd_determination_2} shows the results of this procedure.}

\begin{figure}[htbp!]
\subfigure[\new{Level 1: Lexicographic optimization and one weighted sum (three training runs)}\label{fig:knee2_level1}]
{\includegraphics[width=0.48\textwidth,trim=10 0 37 39]{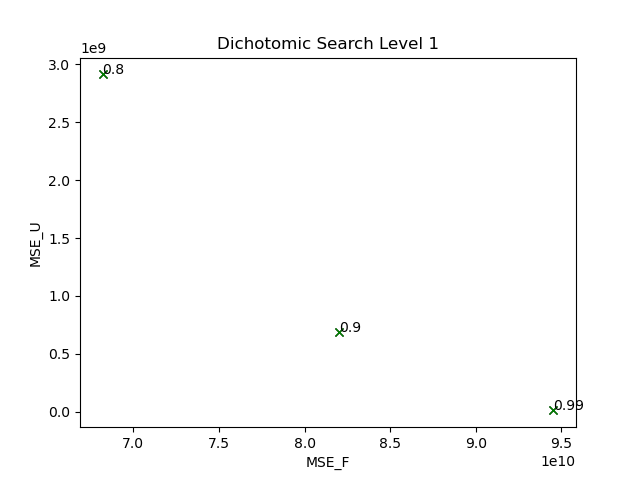}}
\hfill
\subfigure[\new{Level 2: Five training runs, bisection used}\label{fig:knee2_level2}]
{\includegraphics[width=0.48\textwidth,trim=10 0 37 39]{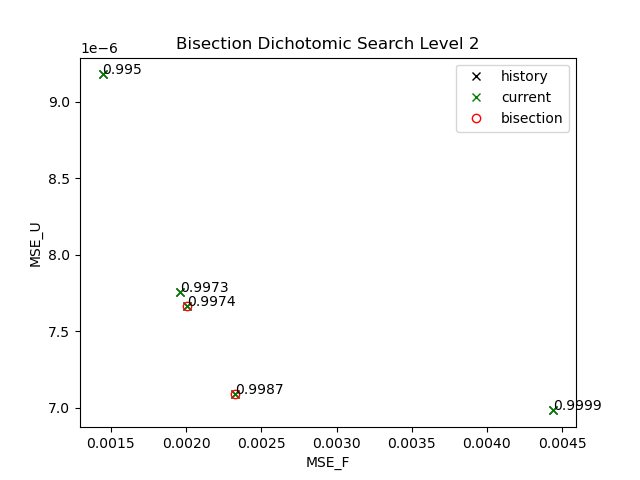}}
\vspace{0.5cm}
\\
\subfigure[\new{Level 3: Nine training runs} \label{fig:knee2_level3}]
{\includegraphics[width=0.48\textwidth,trim=6 0 37 39]{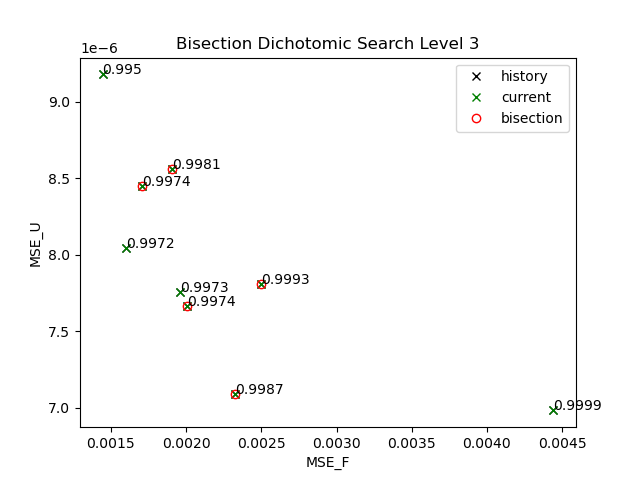}}
\caption{\new{Pareto front approximation using dichotomic search with a starting learning rate of 0.003 and learning rate schedule with starting weights $\alpha_1 = 0.995$ and $\alpha_2 = 0.9999$. 
Favorable trade-offs can be identified starting from level~3.}\label{fig:knee2}}
\end{figure}

\new{We choose $\beta=0.00000001314$ and $\kappa=0.001$. 
With these values, we start another dichotomic search to find a loss weighting with a good trade-off. 
Since the underlying physical system in this case fits the data less well than in the previous short-term prediction (since the long-term data consist of multiple waves with different infection peaks), 
we chose a search window closer to $\alpha=1$ with $\alpha=[0.995,0.9999]$ to weight the data loss even higher. 
The results can be seen in Figure~\ref{fig:knee2}.}

\begin{figure}[htpb!] \centering 
\includegraphics[width=1.0\textwidth]{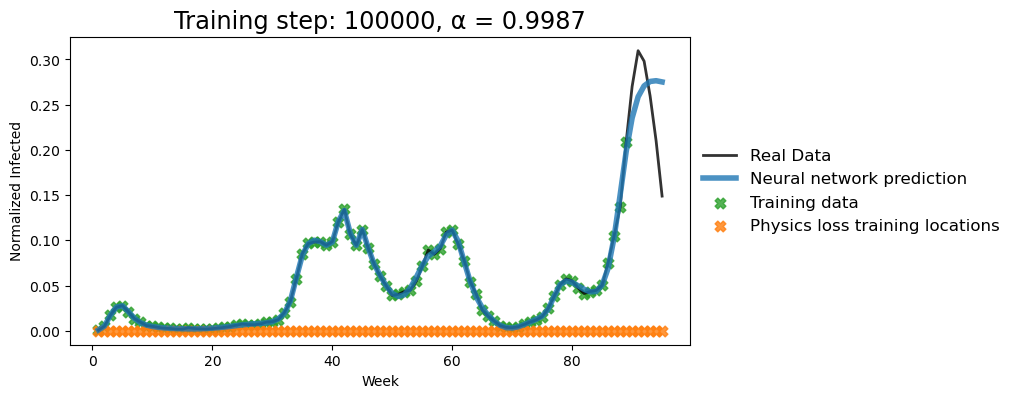}  
\caption{\new{Infection numbers obtained from the reported data (black) or the training of the PINN with the loss terms $\MSEU$ and $\MSEF$ (blue) from the 1$^{st}$ considered week to the 95$^{th}$ week for $\alpha=0.9987$. 
The training data set covers the calendar weeks 10 in 2020 to 45 in 2021 ($1^{st}$ to $89^{th}$ considered week).}}
\label{fig:validlongdata1}
\end{figure}

\new{We identify the weighting $\alpha=0.9987$ as a good trade-off solution that does not lose much in the $\MSEU$-loss and is much better than $\alpha=0.9999$ in the $\MSEF$-loss. 
We decide to use this weighting for our long-term prediction. 
The results can be seen in Figure~\ref{fig:validlongdata1}.}

\new{We note that the quality of the prediction is lower compared to the short-term prediction.
 This is not surprising given that the underlying SVIHR system is designed to simulate one wave of infection. 
To improve these results, time-varying functions for  vaccination and transmission rates would need to be included, which we plan to investigate in future work.}

%%%%%%%%%%%%%%%%%%%%%%%%%%%%%%%%%%%%%%%%%%%%%%%%%%%%%%%%%%%%%%%%%%
\section{Conclusion and Outlook}\label{sec:conclusion} 
We consider \textit{physics-informed neural networks} (PINNs), a Deep Learning technique that combines data and \new{physical knowledge.  
The predictions for COVID-19 infection rates are used as a case study. Our main contribution is a new perspective on the trade-off between data loss and residual loss in PINN training. 
We present an inherently biobjective method that efficiently identifies near-ideal knee solutions. 
This is complemented by an inverse modeling approach to determine the model parameters that govern the dynamics of the system, based on a numerical solution of the associated system of differential equations.} 

In our approach, the data loss \new{is computed for data on COVID-19 infection rates in Germany. The PINN predicts} the sizes of the compartments of an established \textit{susceptible-vaccinated-infected-hospitalized-recovered} (SVIHR) model, \new{with a focus on the compartment of infected individuals.} 
The residual loss is derived from a system of ODEs based on the proposed SVIHR model, which mathematically describes the dynamics of transitions between different compartments and infectious disease transmission in a population affected by the COVID-19 pandemic. 
We propose an NSFD scheme especially designed for the numerical solution of the SVIHR model. 
Its solution \new{is used to estimate the transmission risk $\beta$ and the residual transmission probability after vaccination $\kappa$ that govern the dynamics.}

\new{Our results show that the prediction quality of the PINN is highly dependent on the considered  data, and on the weighing \new{parameter $\alpha$ that combines the two conflicting} loss functions \new{in a weighted sum} $\mathcal{L}_\alpha$. 
A combination of short-term data and an optimized choice of the weighting parameter $\alpha$ leads to a very good prediction of the first omicron wave.}

The approach is based on a dichotomic search method that iteratively approximates the Pareto front and \new{identifies near-ideal knee} solutions (i.e., trained networks) with comparatively few training runs. 
We found that the preferred values of the weighting parameter $\alpha$ in the total loss function were greater than \new{0.99} in most cases, 
thus giving greater weight to data loss than to residual loss. 
\new{This numerical value has to be interpreted with care due to the different scalings of the data loss and the residual loss.} 
We emphasize that the biobjective approach to PINN training can be extended to PINNs regardless of the number of loss terms or application field. 
\new{While appropriate weighting parameters $\alpha$ could theoretically be obtained by scanning the weights of reasonable candidate values, this would require a prohibitively large number of individual neural network trainings, growing exponentially with the number of loss functions.} 

\new{Finally, we use NSFD parameter estimation and dichotomic search to perform long-term prediction for COVID-19 infection rates, using most of the available data as training data for the delta wave prediction. 
We find that the PINN provides reasonable results even for long-term predictions. 
However, since the dynamics in the long term are affected by many different aspects and measures, the physical model is even less accurate in this case, which explains the slightly worse performance compared to the short term predictions.}

In future work, time-varying functions for vaccination and transmission rates will be included\new{, i.e.\ $\beta=\beta(t)$, $\kappa=\kappa(t)$,} to account for seasonal and variant-dependent fluctuations, cf.\ \cite{jagan}. 
Time-variability in the transmission rate should especially be taken into consideration with respect to long-term predictions, 
in which the training data set covers multiple months and thus includes different %experienced 
mutations, lockdown measures, and seasons.
\new{To this end, it is also necessary to incorporate a reflux from the recovered compartment $R$ to the susceptible compartment $S$.}
It is worth noting that we only dealt with two loss terms here and therefore used a biobjective optimization model. 
The concept of multiobjective PINN training can easily be extended to consider more than two objective functions. 
This occurs in more advanced applications such as solving PDEs by PINNs, e.g., using gradient-enhanced PINNs (gPINNs) \cite{Yu21}, in which case the number of loss terms is greater than two and a multiobjective approach is required. 
For example, in \cite[Section~2.2]{Yu21}, the number of loss terms is $3+d$, where $d$ is the dimension of the spatial domain.
Here, one has to identify similar and conflicting training objectives in order to avoid an overly large parameter set. %decide which terms are combined into a common term and which terms become a weighting factor. 
This extension to  multiobjective optimization approaches will be the content of a follow-up article.

%

%%%%%%%%%%%%%%%%%%%%%%%%%%%%%%%%%%%%%%%%%%%%%
%\section*{Acknowledgements}  
%This research was supported ....

\nocite{nagel2022}  % preprint Oct 2022
\nocite{yang2021b}
%%%%%%%%%%%%%%%%%%%%%%%%%%%%%%%%%%%%%%%%%%%%%%
%\section*{References}

\bibliography{PINN}

\end{document}